%% file: main.tex
\pdfoutput=1
\documentclass[11pt, a4paper, logo]{arxivreport}

\usepackage[authoryear, sort&compress, round]{natbib}

\input{math_commands.tex}

\definecolor{citeblue}{RGB}{39,115,219}
\hypersetup{
  colorlinks,
  citecolor=citeblue,
  linkcolor=red,
  urlcolor=blue}
\usepackage{booktabs}
\usepackage{nicefrac}
\usepackage{wrapfig}
\usepackage{paralist}
\usepackage{listings}
\usepackage{upquote}
\usepackage[capitalize,noabbrev]{cleveref}

\lstdefinestyle{skilldump}{
  basicstyle=\ttfamily\scriptsize,
  breaklines=true,
  breakautoindent=false,
  breakindent=0pt,
  columns=fullflexible,
  keepspaces=true,
  showstringspaces=false,
  upquote=true,
  aboveskip=0pt,
  belowskip=0pt,
}

\lstdefinestyle{inlinecode}{basicstyle=\ttfamily,upquote=true,columns=fullflexible,keepspaces=true}
\newcommand{\code}[1]{\lstinline[style=inlinecode]|#1|}

\usepackage[framemethod=TikZ]{mdframed}

\definecolor{ExampleBg}{HTML}{ffffff}
\definecolor{ExampleTitle}{HTML}{629677}
\newcounter{example}

\mdfdefinestyle{examplestyle}{
    roundcorner=5pt,
    backgroundcolor=ExampleBg,
    linecolor=ExampleTitle,
    outerlinewidth=0.3pt,
    frametitlebackgroundcolor=ExampleTitle,
    frametitlefont={\small\sffamily\bfseries\color{white}},
    skipabove=10pt,
    frametitleaboveskip=3pt,
    frametitlebelowskip=3pt,
}

\newenvironment{example*}[1][]{\ifstrempty{#1}{\mdframed[style=examplestyle]}{\mdframed[style=examplestyle, frametitle={#1}]}}{\endmdframed
}
\crefname{example}{Example}{Examples}

\definecolor{TakeAwayBg}{HTML}{ffffff}
\definecolor{TakeAwayTitle}{HTML}{364156}
\newcounter{takeaway}
\mdfdefinestyle{takeawaystyle}{
    roundcorner=5pt,
    backgroundcolor=TakeAwayBg,
    linecolor=TakeAwayTitle,
    outerlinewidth=0.3pt,
    frametitlebackgroundcolor=TakeAwayTitle,
    frametitlefont={\sffamily\bfseries\color{white}},
    skipabove=10pt,
    frametitleaboveskip=3pt,
    frametitlebelowskip=3pt,
}

\crefname{takeaway}{Takeaway}{Takeaways}

\newcommand{\method}[0]{\textsc{TeleTune}}
\newcommand{\std}[1]{${\scriptstyle\,\pm#1}$}

\title{\method{}: Evolving Agent Skills\\From Offline Telemetry}
\reportdate{Oct 2026}

\author[1,$\ast$]{Justin Chih-Yao Chen}
\author[2]{Elias Stengel-Eskin}
\author[2]{Yan Chen}
\author[2]{Pol Llado}
\author[2]{Scott Counts}
\author[1]{Mohit Bansal}
\author[2]{\\Benjamin Van Durme}
\author[2]{Harsh Jhamtani}
\author[2]{Gaurav Verma}

\affil[1]{UNC Chapel Hill}
\affil[2]{Microsoft}

\codeurl{https://microsoft-teletune.github.io/}
\correspondence{\href{cychen@cs.unc.edu}{cychen@cs.unc.edu}, \href{gauravverma@microsoft.com}{gauravverma@microsoft.com}}

\hypersetup{pdftitle={TeleTune: Evolving Agent Skills From Offline Telemetry}}

\begin{document}

\maketitle
{\renewcommand{\thefootnote}{\fnsymbol{footnote}}\footnotetext[1]{Work done while interning at Microsoft.}}
\begin{abstract}
Computer-use agents need to capture procedural knowledge of how people use software.
User telemetry, i.e., logs of interface states and user actions, offers a scalable source of this knowledge.
However, learning reusable skills from these logs requires addressing three challenges: \textbf{(1) Goal Underspecification}, since logs do not record the goal behind each action; \textbf{(2) Non-Replayability}, since past activity cannot be replayed to evaluate skill updates; and \textbf{(3) Interleaved Trajectories}, since logs may mix several tasks without marking their boundaries.
To address these challenges, we introduce \method{}, a framework for learning a textual skill library from offline logs without recorded goals, cannot be replayed during optimization, and may interleave tasks.
The library contains workflows describing cross-instance procedures and skills guiding individual task decisions, both provided as context to the agent at test time.
\method{} uses action-prediction errors on logged trajectories to propose library edits and keep only those that improve held-out action-prediction accuracy, which we call \emph{skill-guided progress}.
The learned workflows also enable retrieval of demonstrations that cover the subgoals of a new task.
At test time, the agent is provided with the learned library and the workflow-based retrieved demonstrations.
Experiments on WorkArena and Online-Mind2Web show that \method{} outperforms random retrieval, Agent Workflow Memory (AWM), and their combination. We find that the best baseline varies by setting, whereas \method{} achieves average success rates of $77.1\%$ and $80.6\%$, respectively, improving over the strongest baseline on each benchmark by $6.7\%$ and $7.7\%$.
Under the heaviest perturbation of the WorkArena training data, \method{} keeps the highest average success rate at $68.5\%$, $6.3\%$ above the strongest baseline.
Finally, our analyses show that skill optimization and workflow-based retrieval are complementary, that optimizing on fixed logs costs 5 to 75 times fewer tokens than validating the same edits with live episodes, and that skill-guided progress tracks the live success rate.
\end{abstract}

\section{Introduction}
\label{sec:introduction}

\begin{figure*}[t!]
    \centering
    \includegraphics[width=\linewidth]{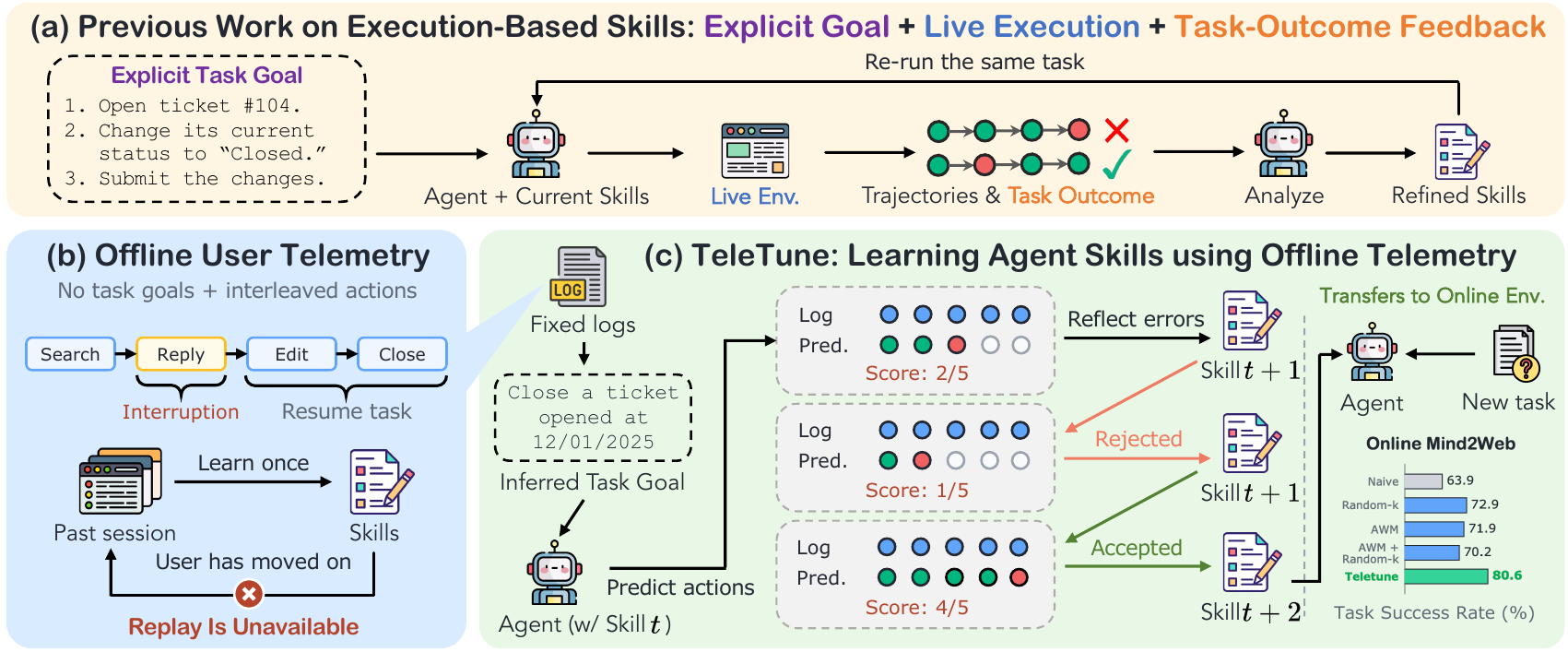}
    \vspace{-1.5em}
    \caption{\textbf{Learning Transferable Agent Skills From Offline User Telemetry.}
(a) Execution-based skill refinement uses explicit task goals and outcome feedback with a live environment.
(b) User telemetry does not record the goal behind each action, and revising a skill cannot produce new outcomes for the logged activity. The logs may also interleave tasks.
(c) \method{} infers a goal for each logged trajectory, uses prediction errors on the fixed logs to propose skill edits, and keeps edits that improve skill-guided progress. The learned library is then evaluated on unseen tasks.
}
    \label{fig:intro}
\end{figure*}

Completing a task in an unfamiliar software system requires knowing more than which buttons to click. For example, an agent handling a service request may need to locate the relevant record, check its fields, and follow a local procedure before changing its status. Learning reusable skills from experience offers a way to provide agents with such procedural knowledge \citep{wang2024awm,ouyang2025reasoningbank,shinn2023reflexion,wang2023voyager,zhao2024expel,zheng2025skillweaver}. For agents adapting to an organization's tools and practices, one source of this experience is \emph{user telemetry}, the logs of interface states and actions recorded as people use software. These logs capture how users carry out recurring procedures and could help an agent complete similar tasks in the future. However, these logs record ongoing user activity whose goals and task boundaries are implicit, unlike the individual, goal-labeled trajectories used in the skill-learning literature.
Learning from this ongoing activity raises three related challenges visualized in \cref{fig:intro}(a) and (b):
\textbf{(1) Goal Underspecification:} Logs record what a user did, but typically do not record the goal behind each action sequence. The agent must learn from actions whose purpose is implicit in the interface state and preceding history.
\textbf{(2) Non-Replayability:} Telemetry records actions taken by real users while completing their own tasks. We cannot expect users to repeat those tasks each time we update a skill library just to help validate it. Nor do logs necessarily preserve a resettable environment in which an agent can try the task with the updated library. We must therefore evaluate skill updates offline from the recorded activity, without the fresh outcome feedback available through live interaction.
\textbf{(3) Interleaved Trajectories:} Logs may interleave several tasks without marking their boundaries. A user might interrupt a record update to look up information elsewhere, handle an unrelated request, and then return to the original record.

Prior work addresses parts of this setting. AWM and ReasoningBank learn reusable guidance from goal-labeled experience \citep{wang2024awm,ouyang2025reasoningbank}, while NNetNav and robotic process mining address goal-free exploration and unsegmented logs, respectively \citep{murty2024nnetnav,leno2020unsegmented,leno2021rpm}. None of these approaches addresses all three challenges together, which motivates our question:
\textit{can an agent learn reusable skills from logs that carry no goals, cannot be replayed to evaluate updates, and interleave several tasks?} We illustrate this setting in \cref{fig:intro} (c),
and introduce Telemetry-Informed Tuning for Agent Skills (\method{}), a framework for learning a reusable textual skill library from offline logs without recorded goals or environment access during optimization. Our central idea is to use the prediction of recorded actions as the learning signal for building and revising the skill library. The library contains (1) \emph{workflows}, describing procedures shared across task instances, and (2) \emph{skills}, which guides individual decisions. Both are editable text provided in the prompt of an agent at test time. Like SkillOpt \citep{yang2026skillopt}, we treat the skill text itself as the object of optimization and keep only edits that improve held-out performance. However, SkillOpt learns from individual tasks with known goals and validates each edit by executing tasks in a live environment. Telemetry offers none of these conditions, since its trajectories do not specify goals, cannot be replayed, and may interleave several tasks.

Concretely, \method{} predicts the next action at each step of a logged trajectory, conditioned on the inferred goal, current library, interface state, and preceding action history. After each prediction, we append the recorded action to the history, so an early error does not derail later steps and every scoreable step contributes to the learning signal. We call the resulting action-prediction accuracy \emph{skill-guided progress}. An optimizer examines the first prediction error in each trajectory and proposes additions, revisions, or removals to the entries in the library. The edits form one candidate library, which we keep only if it improves skill-guided progress on held-out trajectories. This loop allows us to refine skills using fixed logs without requesting new outcomes from the environment.
Skill-guided progress can also serve as an \emph{offline proxy for the live success rate}: in our WorkArena L1 run, we find that improvements in this metric track improvements in the live success rate (\cref{fig:optimization}). This lets us compare skill libraries using existing trajectories without a live test environment, which is often unavailable  for real-world systems.
At test time, \method{} uses the learned workflows to retrieve demonstrations that together cover a new task's subgoals. The agent then receives the task goal, the library, and these demonstrations in its prompt and acts in the live environment.

We evaluate \method{} on WorkArena \citep{drouin2024workarena, boisvert2024workarenapp} and Online-Mind2Web \citep{deng2023mind2web,xue2025illusion}. WorkArena tests transfer from successful oracle trajectories to new tasks at the atomic L1 and compositional L2 levels. To isolate the effect of task interleaving, we also construct perturbed training pools with varying interruption rates. To test with human-recorded activity, we learn from annotator-recorded Mind2Web trajectories on real websites \citep{deng2023mind2web} and evaluate on new tasks through live website interaction \citep{xue2025illusion}.
With GPT-5.4 as the agent, \method{} outperforms random demonstration retrieval, Agent Workflow Memory, and their combination on both benchmarks. \method{} reaches $92.3\%$ success on WorkArena L1 and $61.8\%$ on L2, improving over the strongest baseline by $12.0\%$ and $1.4\%$, respectively. On Online-Mind2Web, it also reaches $80.6\%$ with a $7.7\%$ improvement.
Moreover, \method{} shows the highest average WorkArena success rate under the heaviest controlled perturbation, reaching $68.5\%$, $6.3\%$ above the strongest baseline. Our analyses further show that (1) skill optimization and workflow-based retrieval are complementary, (2) optimizing on fixed logs costs 5 to 75 times fewer tokens than validating the same edits with live episodes, and (3) skill-guided progress tracks the live success rate during optimization. On WorkArena L1, the live success rate rises from $74.1\%$ to $92.3\%$ as skill-guided progress improves, which suggests that skill-guided progress can serve as an offline proxy for evaluating skill updates.
\section{\method{}: Telemetry-Informed Tuning for Agent Skills}
\label{sec:method}
\begin{figure*}
    \centering
    \includegraphics[width=\linewidth]{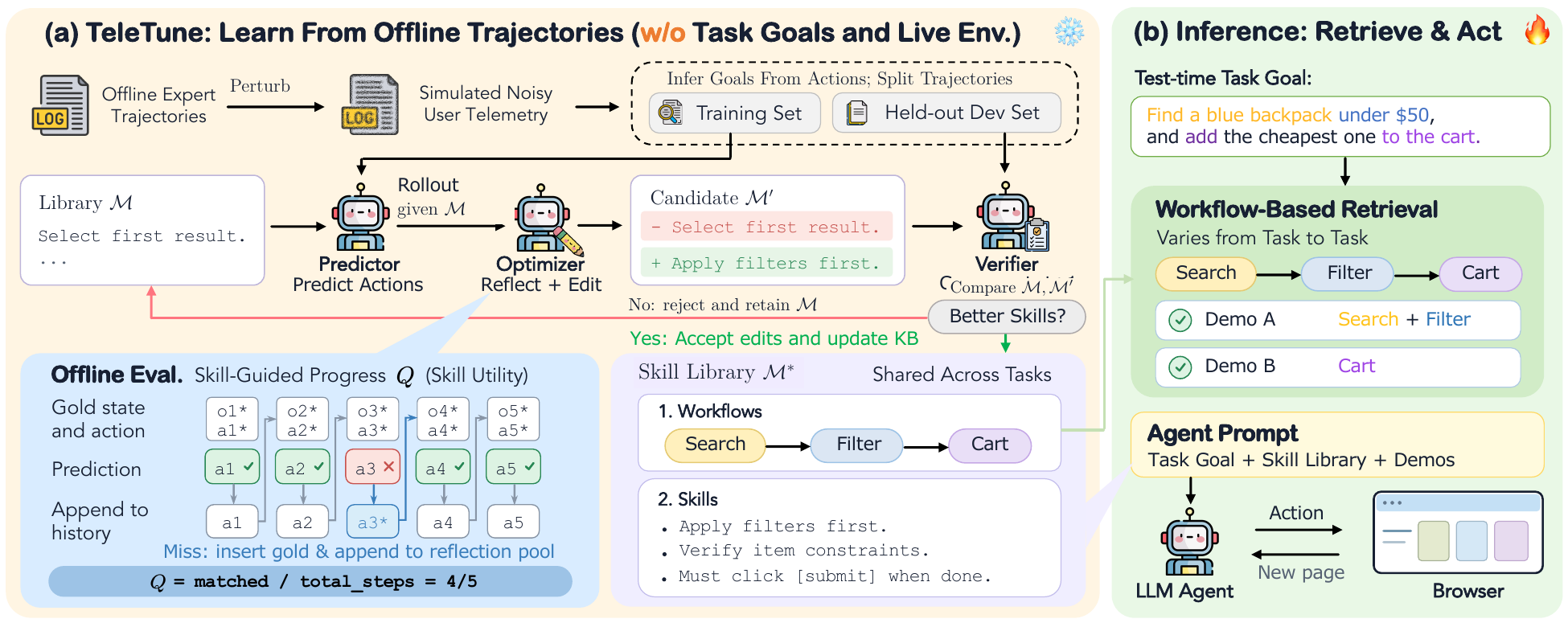}
    \vspace{-1.5em}
    \caption{\textbf{Overview of \method{}.}
(a) Training trajectories are split into train and dev splits. The predictor runs through each train-split trajectory with the current skill library $\mathcal{M}$ in its prompt. After each prediction, we append the recorded action to the history (teacher forcing), so an error does not derail later steps. The optimizer reflects on the first prediction error in each trajectory and proposes edits, giving a candidate library, and we keep the one with higher skill-guided progress. (b) At inference, workflow-based retrieval selects complementary demonstrations for the new task. The task goal, skill library, and retrieved demonstrations form the agent prompt.}
    \label{fig:method}
\end{figure*}

\subsection{Problem Formulation}
\label{sec:problem}
We model a computer-use task as a partially observable Markov decision process, with a reward defined for a task goal $g$. The agent never observes the state directly. At step $t$ it sees an observation $o_t$, the current browser page in our experiments, and keeps an action history $h_t=(a_1,\ldots,a_{t-1})$. It acts with a frozen language-model policy $\pi_{\theta}$. The object we learn is the skill library $\mathcal{M}$, a text document of workflows and skills (\cref{sec:skilltune}) placed in the agent's prompt and shared across tasks.
During skill optimization, no recorded goal is available, and no demonstrations are retrieved. We call the policy the \emph{predictor} in this phase, and we give it an inferred goal $\hat{g}$ written from the recorded actions (see \cref{app:goal-inference} for the prompt), so the train-time policy is $\pi_{\theta}(\cdot\mid \hat{g},o_t,h_t,\mathcal{M})$.
At test time, we condition the agent on the task goal $g$ and on a set of retrieved demonstrations $\mathcal{E}$ (\cref{sec:workflow-retrieval}),
and it predicts the next action:
\begin{equation}
\label{eq:policy}
    a_t \sim \pi_{\theta}
    \left(
        \cdot \mid g,o_t,h_t,\mathcal{M},\mathcal{E}
    \right).
\end{equation}
Our goal is to improve the agent by learning $\mathcal{M}$ and constructing $\mathcal{E}$ while leaving $\theta$ unchanged.

\noindent\textbf{Training Trajectories.}
We assume a collection $\mathcal{D}$ of successful trajectories recorded before optimization. Each trajectory $\tau=((o_1,a_1),\ldots,(o_{|\tau|},a_{|\tau|}))$ is a sequence of interface states and the actions taken in them. We assume that each trajectory holds only states and actions, with no task goal, and that the logs cannot be replayed, so \method{} never interacts with the training environment.
We split $\mathcal{D}$ into a train split and a dev split. We use the train split to find the prediction errors that motivate edits and the dev split to decide whether to accept them.

\noindent\textbf{Offline Evaluation by Teacher Forcing.}
We run the predictor step by step through each trajectory with the library in its prompt (\cref{fig:method}a). At step $t$, it sees the inferred goal, the history, and on Mind2Web the recorded page $o_t$, and predicts the next action $\hat{a}_t$. We append $\hat{a}_t$ to the history if it matches the recorded action $a_t$ and append $a_t$ otherwise (i.e., teacher forcing). An early error therefore does not derail the rest of the trajectory, and every later step still gives feedback.
We score only the actions that are predictable from the recorded context. Actions whose target is not a visible label, such as a CSS selector or a positional index like \code{nth=0}, are kept in the history but are not scored, and a prediction counts as correct when its action type matches and its target and value match closely (\cref{app:hyperparameters}).
We call the per-trajectory action-prediction accuracy \emph{skill-guided progress},
\begin{equation}
    Q(\tau,\mathcal{M})
    =
    \frac{1}{|S_\tau|}
    \sum_{t\in S_\tau}
    \mathbf{1}\left[\hat{a}_t=a_t\right],
\end{equation}
where $S_\tau$ is the set of scoreable steps in $\tau$.

\subsection{Skill Optimization}
\label{sec:skilltune}

The skill library $\mathcal{M}$ is one structured text document that is inserted directly into the agent's context. It contains two kinds of entries. A \emph{workflow} gives the ordered steps of a procedure that recurs across tasks, with the values that vary across instances written as placeholders. A \emph{skill} resolves a single decision that a procedure leaves open, such as which field to fill first or whether to enter a value exactly as the task gives it. \cref{app:learned-skills} shows three full libraries containing both.  The library starts with workflows only, induced from the training trajectories with a workflow-induction prompt (\cref{app:workflow-induction}), and with no skills. Skill optimization then repeatedly applies an offline reflect-edit-validate loop (\cref{fig:method}a).
At each optimization step, we collect prediction errors on the train split, ask the optimizer to propose a small set of edits, and keep the edits only if they raise skill-guided progress $Q$ on the dev split.
Optimization runs in two stages. The workflow stage edits only workflows, and the skill stage then edits only skills while the workflows stay fixed. Each stage ends after five consecutive steps without an accepted edit or after 25 steps (\cref{app:hyperparameters}).
The predictor and the optimizer are LLMs with the same backbone as the agent used at inference time.

\noindent\textbf{Reflect.}
Given the current library, we run the predictor through the train-split trajectories as in \cref{sec:problem}.
For each trajectory with at least one error, we record the first prediction error as a failure case, consisting of the inferred goal, the preceding history, the page on Mind2Web, the predicted action, and the recorded action.
The optimizer reads a fixed-size batch of these cases during the Edit step. Each case contains only what the predictor could see. The inferred goal is written from the whole trajectory and can reflect later steps, but its prompt forbids describing the action sequence (\cref{app:goal-inference}), which limits how much the optimizer can learn about future actions from it.

\noindent\textbf{Edit.}
The optimizer reads the failure batch, the current library $\mathcal{M}$, and the list of edits rejected since the last accepted step, and proposes a small set of edits that are applied together to give a candidate library $\mathcal{M}'$.
For workflows, an edit requests that a workflow be distilled, replaced, or removed.
For skills, an edit adds, replaces, or removes one entry, and each new skill is a short rule tied to an observed failure.
Skills are written from a single failure case, while workflows are distilled from several trajectories.
Restricting the optimizer to explicit edit operations keeps each change readable and prevents it from rewriting the whole library.
For example, GPT-5.4 accepted the following skill on WorkArena L1: \emph{``In list filters, if the needed value is absent or no value widget is available after setting field and operator, finish the filter by running it instead of guessing a value.''} The optimizer wrote it from a single trajectory in which the recorded user left the \emph{assigned to} condition empty and clicked \emph{Run}, while the predictor filled in a user name. \cref{app:learned-skills} shows full libraries.

\noindent\textbf{Validate.}
Once per step, after the proposal, we score the candidate and current libraries on the same dev-split trajectories with skill-guided progress $Q$ (\cref{sec:problem}). The candidate library $\mathcal{M}'$ replaces $\mathcal{M}$ only if its mean $Q$ over the dev split is higher; otherwise we keep $\mathcal{M}$.
We checkpoint accepted libraries and record rejected edits. We include the rejected edits in the optimizer's input in later steps, so it does not propose the same change again. In summary, reflection proposes changes from observed failures, validation keeps only those that transfer to the dev split, and the overall pipeline does not require an online environment.

\subsection{Workflow-Based Retrieval}
\label{sec:workflow-retrieval}
Workflow-based retrieval gives the agent demonstrations that cover the subgoals a new task requires. Learned abstractions are often not enough on their own, since the agent also needs to see the contexts in which they are used \citep{zheng2024synapse,stengeleskin2024regal,wang2024awm}.
Similarity-based retrieval is one option, but on compositional tasks it often returns several demonstrations of the same operation. In our preliminary experiment, with $k{=}8$, it performs similarly to Random-$k$ on WorkArena L1 ($+2.1\%$) but falls below it on L2 ($-4.5\%$).
We instead use the optimized workflows $\mathcal{W}^{*}$ as an intermediate representation between task goals and trajectories. Offline, each training trajectory is annotated with the workflows it instantiates, and common workflows are down-weighted so that generic operations do not dominate.
For a new task, a planner decomposes the task goal into subgoals and maps each subgoal to candidate workflows. A training trajectory covers a subgoal when the workflows it instantiates include one the planner matched to that subgoal, and the match clears a confidence threshold. A greedy selector then repeatedly picks the trajectory that covers the most still-uncovered subgoals, until every subgoal is covered, no confident match remains, or the demonstration budget is reached.
The greedy selector thus favors complementary demonstrations over redundant neighbors. \cref{app:retrieval} gives the weighting, the relevance score, and the selection rule; \cref{tab:skilltune-ablation} compares this selector with random retrieval.

\subsection{Inference}
\label{sec:inference}

After optimization, the final skill library $\mathcal{M}^{*}$ is fixed and shared across all test tasks. For each task goal $g$, retrieval builds $\mathcal{E}$ once before interaction begins, and the prompt then holds the task goal, the library, and the retrieved demonstrations (\cref{fig:method}b). The task goal $g$ is available to both the retrieval planner and the agent, while the retrieved demonstrations carry no goals of their own. At each step, the agent predicts an action with \cref{eq:policy}, executes it, observes the resulting page, and extends its history, while the library and demonstrations stay fixed.
\method{} therefore adapts at two levels. Skill optimization learns global knowledge that transfers across tasks, and workflow-based retrieval assembles local evidence to cover subgoals of the current task.
\section{Experimental Setup}
\label{sec:setup}
\noindent\textbf{Dataset.} We evaluate \method{} on two datasets. \textbf{(1) WorkArena:} a suite of knowledge-work tasks on a live ServiceNow instance~\citep{drouin2024workarena}. L1 tasks are generally easier and consist of atomic operations such as filling a form.
L2 tasks are from WorkArena++ \citep{boisvert2024workarenapp}, compose several such operations under one instruction and require the agent to carry information between them.
We prune task families whose oracle is unsuitable for learning or evaluation: those with empty action traces, whose expert policy fails because of broken ServiceNow fixtures or controls, and L2 tasks whose solution is to report infeasibility.
This leaves 270 training and 270 test instances on L1, and 216 training and 198 test instances on L2.
\textbf{(2) Mind2Web:} we train on annotator-recorded trajectories from the original Mind2Web training set \citep{deng2023mind2web} and evaluate through live interaction on Online-Mind2Web \citep{xue2025illusion}. To align the training and test domains and to reduce cost, we keep only test tasks whose website category also appears in the original Mind2Web training set, and we downsample the training trajectories to follow the test category distribution (details in \cref{app:data}). This yields 640 training trajectories and 122 test tasks. \cref{tab:data-stats} provides more detailed statistics on both datasets.
We would like to highlight that Online-Mind2Web matches our setting directly. The training data are trajectories recorded on real websites in the past, and evaluation asks the agent to complete new tasks on live sites, with no guarantee that the recorded trajectories can be replayed since the sites may have changed. All training data are \textbf{goal-free}, i.e., only at test time, the agent will receive a task goal.

\noindent\textbf{Models.} We use GPT-5.4 for the main results
and Qwen3.6-35B-A3B-FP8 and Mistral-Medium-3.5-128B for the cross-model analysis.\footnote{Hereafter, we refer to them as GPT-5.4, Qwen-3.6, and Mistral-3.5, respectively.} Each backbone learns its own skill library. The same backbone infers the goals, induces the workflows, optimizes the skills, and acts as the agent at test time, so an agent is always paired with the library learned by its own backbone.

\noindent\textbf{Baselines.} \emph{Naive} runs the agent with the task goal and uses no training trajectories. \emph{Random-$k$} adds $k=8$ randomly sampled training demonstrations, which carry no goals. \emph{AWM} \citep{wang2024awm} induces workflows from recurring subroutines in the training trajectories and places them in the prompt. \emph{AWM $+$ Random-$k$} combines the two. In our workflow-based retrieval, we also use $k=8$ training demonstrations. All methods share the same model, test tasks, and step budget. As with \method{}, AWM induces its workflows with the agent's own backbone.

\noindent\textbf{Evaluation Protocol.}
Every method runs on the same test tasks with three seeds. On WorkArena, the maximum step budget is set to 100 steps for L1 and 200 for L2.
On Online-Mind2Web, the step budget is set to 200. Following \citet{xue2025illusion}, we judge the completed trajectories with \textsc{WebJudge} using \texttt{o4-mini}. \textsc{WebJudge} extracts the instruction's requirements, selects relevant screenshots, and returns a binary verdict for the task.
Live websites sometimes block automated browsers with a CAPTCHA or a bot-detection page, which stops the agent from completing the task. On Online-Mind2Web, we therefore report a discounted success rate, which excludes episodes that hit such a page and computes success over the remaining episodes.
\section{Results and Analysis}
\label{sec:results}

\subsection{Main Results}
\label{sec:main-results}
\begin{table}[t]
\centering
\caption{Comparison of \method{} with retrieval and workflow-induction baselines, using GPT-5.4 as the agent. \method{} achieves the best success rate in every column. It improves over the strongest baseline by $6.7\%$ on WorkArena average and by $7.7\%$ on Online-Mind2Web average, with the largest gains on WorkArena L1 and on the hardest Online-Mind2Web tasks. Results are success rates (\%) averaged over three seeds. \textbf{Bold} marks the best result in each column.}
\vspace{0.5em}
\label{tab:main-results}
\small
\renewcommand{\arraystretch}{1.25}
\begin{tabular}{l ccc cccc}
\toprule
& \multicolumn{3}{c}{\textbf{WorkArena}} & \multicolumn{4}{c}{\textbf{Online-Mind2Web}} \\
\cmidrule(lr){2-4} \cmidrule(lr){5-8}
\textbf{Method} & L1 & L2 & Avg. & Easy & Medium & Hard & Avg. \\
\midrule
Naive
  & 45.6\std{0.6} & 48.0\std{2.0} & 46.8\std{0.9}
  & 78.5\std{6.0} & 66.7\std{10.8} & 48.7\std{1.2} & 63.9\std{4.6} \\
Random-$k$
  & 65.1\std{2.3} & 54.2\std{2.1} & 59.7\std{2.1}
  & 87.3\std{7.4} & 71.3\std{1.7} & 64.7\std{9.4} & 72.9\std{4.0} \\
AWM
  & 75.7\std{1.7} & 51.9\std{2.9} & 63.8\std{1.8}
  & 88.8\std{5.4} & 73.4\std{3.9} & 56.7\std{5.8} & 71.9\std{3.0} \\
AWM $+$ Random-$k$
  & 80.3\std{1.1} & 60.4\std{3.4} & 70.4\std{1.7}
  & 85.6\std{4.6} & 66.5\std{12.0} & 64.0\std{5.7} & 70.2\std{4.4} \\
\method{}
  & \textbf{92.3}\std{2.0} & \textbf{61.8}\std{2.8} & \textbf{77.1}\std{2.1}
  & \textbf{94.5}\std{5.3} & \textbf{76.6}\std{9.7} & \textbf{75.3}\std{2.8} & \textbf{80.6}\std{4.1} \\
\bottomrule
\end{tabular}
\end{table}

\noindent\textbf{\method{} outperforms all baselines on both benchmarks.}
We first compare \method{} with the baselines using GPT-5.4 as the agent (\cref{tab:main-results}). \method{} achieves the best success rate across different levels and benchmarks. On WorkArena, it reaches $77.1\%$ on average, $6.7\%$ above the strongest baseline (AWM $+$ Random-$k$), and shows the largest gain on L1 ($80.3\% \rightarrow 92.3\%$). On Online-Mind2Web, it reaches $80.6\%$, $7.7\%$ above Random-$k$, and the largest gain is on hard tasks ($64.7\% \rightarrow 75.3\%$). Combining the two baselines does not always help. AWM $+$ Random-$k$ is the strongest baseline on WorkArena, yet on Online-Mind2Web it falls below both AWM and Random-$k$ ($70.2\%$ versus $71.9\%$ and $72.9\%$).
On WorkArena L2, \method{} leads AWM $+$ Random-$k$ by $1.4\%$, a smaller margin than on L1. We propose two reasons for this.
First, fewer optimization steps were accepted on L2 than on L1 (4 versus 14, \cref{tab:skill-opt-cost}). L2 has 216 training trajectories spread over 72 task templates, about $3$ per template against $30$ on L1 (\cref{tab:data-stats}), so an edit written from one failure has few similar dev trajectories on which to raise skill-guided progress. The same sparsity may explain why similarity-based retrieval falls below Random-$k$ on L2 but not on L1 (\cref{sec:workflow-retrieval}). With few demonstrations per template, the nearest neighbors of an L2 task tend to repeat one operation, while random samples cover more task families. Second, L2 require about three times as many steps as L1 tasks (\cref{tab:step-counts}) and require carrying information between operations. Thus, adding skills for individual decisions may not address the primary bottlenecks for L2 tasks.

\begin{wraptable}{r}{0.44\textwidth}
\vspace{-1.7em}
\centering
\caption{\method{} generalizes to Qwen-3.6 and Mistral-3.5. Each backbone learns its own skill library and uses it at test time.}
\label{tab:cross-model}
\medskip
\small
\setlength{\tabcolsep}{5pt}
\renewcommand{\arraystretch}{1.25}
\resizebox{\linewidth}{!}{\begin{tabular}{l cc cc}
\toprule
& \multicolumn{2}{c}{\textbf{Qwen-3.6}} & \multicolumn{2}{c}{\textbf{Mistral-3.5}} \\
\cmidrule(lr){2-3} \cmidrule(lr){4-5}
\textbf{Method} & L1 & L2 & L1 & L2 \\
\midrule
Naive
  & 30.4 & 36.4
  & 39.6 & 31.3 \\
Random-$k$
  & 55.9 & 43.9
  & 64.4 & \textbf{38.9} \\
AWM
  & 60.4 & 43.9
  & 74.7 & 38.0\\
AWM $+$ Random-$k$
  & 63.7 & 44.9
  & 72.2 & 38.4 \\
\method{}
  & \textbf{75.6} & \textbf{50.0}
  & \textbf{75.2} & 37.8 \\
\bottomrule
\end{tabular}}
\end{wraptable}

\noindent\textbf{\method{} generalizes to other backbones.}
Next, we test whether \method{} also helps agents built on other backbones (\cref{tab:cross-model}). For each backbone, we rerun the full pipeline, so Qwen-3.6 and Mistral-3.5 each learn their own library from the same training trajectories and use it at test time. With Qwen-3.6 as the agent, \method{} is again the best method on both WorkArena levels. It reaches $75.6\%$ on L1 and $50.0\%$ on L2, which is $11.9\%$ and $5.1\%$ above AWM $+$ Random-$k$. With Mistral-3.5, \method{} is on par with the strongest baseline. It leads AWM by $0.5\%$ on L1 and falls behind Random-$k$ by $1.1\%$ on L2. Overall, \method{} generalizes to other backbones when each backbone learns its own library.

\begin{figure*}[t]
    \centering
    \includegraphics[width=\linewidth]{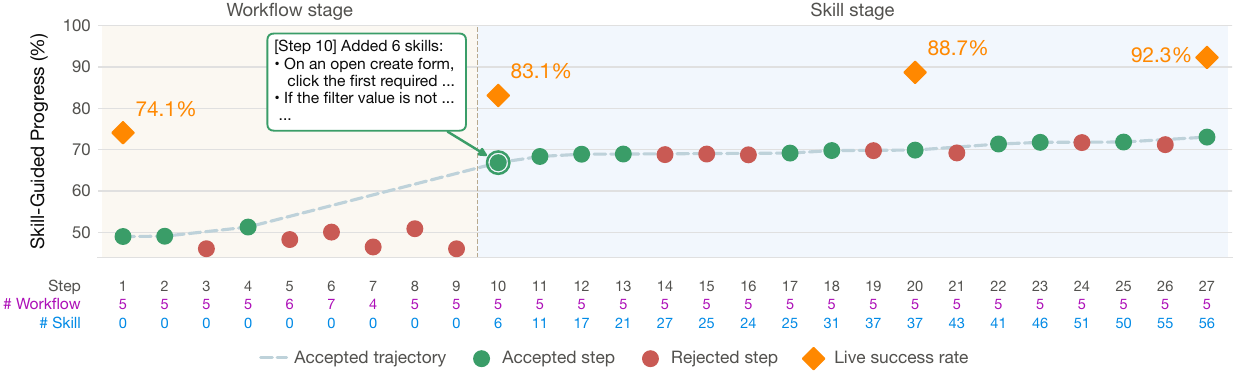}
    \vspace{-1.75em}
    \caption{Skill-guided progress on the dev split during optimization on WorkArena L1 with GPT-5.4. The live success rate rises with skill-guided progress across the four evaluated checkpoints. At step 10, the first skill edit adds six skills, and the skill-guided progress rises from $51.3\%$ to $66.9\%$. We also show the number of workflows and skills in the library at each step (\# workflow and \# skill). We omit the last 5 of the 32 steps, which accept no edits.}
    \label{fig:optimization}
    \vspace{-0.5em}
\end{figure*}

\subsection{Ablations and Analysis}
\label{sec:analysis}

\noindent\textbf{Skill-guided progress is an offline proxy for the live success rate.}
Like prior work, we evaluate every method in live environments. However, many real-world software systems offer no live environment even for evaluation, and logs cannot be replayed. We therefore test whether skill-guided progress on the dev split can stand in for the live success rate, which would make it useful for comparing skill libraries even outside our optimization loop. \cref{fig:optimization} traces the optimization run on WorkArena L1, where we also evaluate four of the accepted libraries live. In the workflow stage (steps 1 to 9), skill-guided progress on the dev split stays near $50\%$. The skill stage starting at step 10 raises skill-guided progress to $73.0\%$, with the largest increase at its first step. Over the same run, the live success rate rises from $74.1\%$ to $83.1\%$, $88.7\%$, and $92.3\%$, in the same order as skill-guided progress. Since the live success rate is never used during optimization, this suggests that skill-guided progress can serve as an offline proxy for it.

\begin{wraptable}{r}{0.44\textwidth}
\vspace{-1.7em}
\centering
\caption{Ablation of skill optimization and workflow-based retrieval with GPT-5.4. Without either component, the agent is the Random-$k$ baseline (first row). The two main components are complementary, as workflow-based retrieval adds $8.6\%$ on L1 with the learned library but $1.7\%$ without it. OM2W is the Online-Mind2Web average.}
\label{tab:skilltune-ablation}
\medskip
\small
\renewcommand{\arraystretch}{1.25}
\resizebox{\linewidth}{!}{\begin{tabular}{cc ccc}
\toprule
\textbf{Skill Opt.} & \textbf{Retrieval} & \textbf{L1} & \textbf{L2} & \textbf{OM2W} \\
\midrule
$\times$      & Random       & 65.1 & 54.2 & 72.9 \\
$\times$      & Workflow     & 66.8 & 52.6 & 74.3 \\
\checkmark    & Random        & 83.7 & 60.8 & 75.1 \\
\checkmark    & Workflow  & \textbf{92.3} & \textbf{61.8} & \textbf{80.6} \\
\bottomrule
\end{tabular}}
\end{wraptable}
\noindent\textbf{Skill optimization and workflow-based retrieval are complementary.}
In \cref{tab:skilltune-ablation}, we toggle the two components of \method{}, skill optimization and workflow-based retrieval. Skill optimization is the larger single factor. With random retrieval, it improves success by $18.6\%$ on L1, $6.6\%$ on L2, and $2.2\%$ on Online-Mind2Web. Workflow-based retrieval does little without the library ($+1.7\%$ on L1, $-1.6\%$ on L2, and $+1.4\%$ on OM2W), which suggests that matched demonstrations transfer poorly on their own. With the learned library, however, workflow-based retrieval adds $8.6\%$, $1.0\%$, and $5.5\%$, respectively, so the two components are complementary. On L1, the full \method{} is $27.2\%$ above the variant with neither component, which exceeds the sum of the two individual gains by $6.9\%$. This also extends to L2 ($+2.6\%$) and OM2W ($+4.1\%$).

\begin{figure*}[t]
    \centering
    \includegraphics[width=\linewidth]{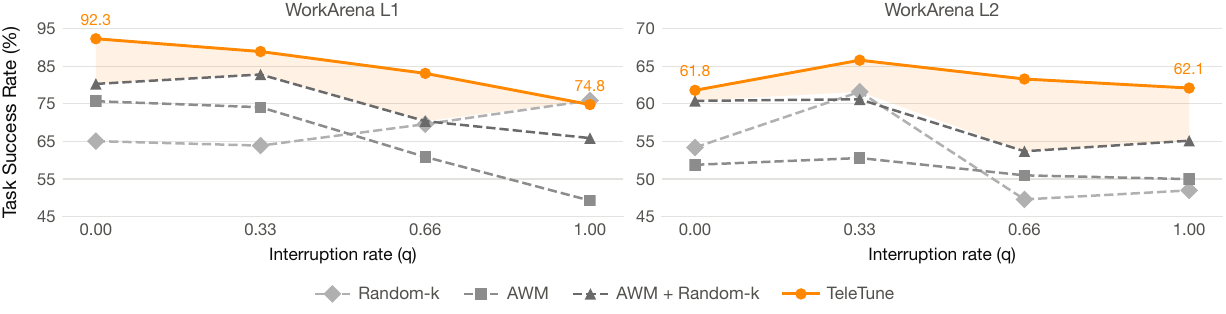}
    \vspace{-2em}
    \caption{\method{} has the highest success rate on both WorkArena levels at every interruption rate $q$ except L1 at $q=1.0$, where Random-$k$ is $1.1\%$ higher. The baselines respond differently as $q$ grows. AWM loses $26.4\%$ on L1, whereas Random-$k$ gains $10.8\%$ on L1 and changes non-monotonically on L2. Naive uses no training pool and is omitted. Expanded values are in \cref{tab:noise-contrast}.
    }
    \label{fig:q_sweep}
    \vspace{-1em}
\end{figure*}

\noindent\textbf{\method{} keeps the highest average success on noisy training pools.}
To test learning from interleaved tasks under controlled conditions, we perturb the WorkArena training pool with an interruption rate $q$ and rebuild every method from the perturbed pool. Specifically, we split each training trajectory into chunks of 2 to 5 actions and, before each chunk, insert with probability $q$ the next chunk of a secondary task drawn from the other training tasks. The number of secondary tasks grows with $q$ ($\lceil 3q \rceil$), and each secondary task advances in order, so a perturbed trajectory switches back and forth between coherent tasks as an interrupted user would. We use $q \in \{0.33, 0.66, 1.0\}$.
\cref{fig:q_sweep} shows success as $q$ grows. \method{} has the highest success on both levels at every $q$ except L1 at $q{=}1.0$, where it is $1.1\%$ below Random-$k$. In contrast, AWM's L1 success falls by $26.4\%$ to $49.3\%$, only $3.7\%$ above the naive agent, which uses no training trajectories at all. We hypothesize that the gap comes from how each method uses the perturbed trajectories. Each perturbed trajectory mixes chunks from up to four tasks, so the same eight demonstrations show the agent more distinct operations than eight clean ones. AWM instead compresses the pool into abstract workflows, which can reduce this diversity.
 \method{} keeps a library entry only if it improves skill-guided progress on the dev split, making it more robust to interleaved tasks.

\begin{wraptable}{r}{0.44\textwidth}
\vspace{-1.7em}
\centering
\caption{Offline cost of skill optimization with GPT-5.4. Steps is the total number of optimization steps. Ckpt.\ reports the number of accepted steps. Token counts (in millions) are split into the optimizer (Opt., edit proposals and workflow induction) and the predictor (Pred., scoring candidates on the dev split), which accounts for almost all of them. L1 and L2 are both from WorkArena, and M2W represents Mind2Web.}
\label{tab:skill-opt-cost}
\medskip
\small
\renewcommand{\arraystretch}{1.25}
\resizebox{\linewidth}{!}{\begin{tabular}{l cc rrr}
\toprule
\textbf{Data} & \textbf{Steps} & \textbf{Ckpt.} & \textbf{Opt.} & \textbf{Pred.} & \textbf{Total} \\
\midrule
L1 & 32 & 14 & 0.24 & 173.60 & 173.84 \\
L2 & 15 &  4 & 0.10 &  23.49 &  23.59 \\
M2W   & 18 &  5 & 0.23 & 329.92 & 330.15 \\
\bottomrule
\end{tabular}}
\end{wraptable}
\noindent\textbf{Optimization is a one-time, offline cost.}
\cref{tab:skill-opt-cost} reports the cost of learning the library. We pay this cost only once, since we learn one library per setting and reuse it unchanged across every test task and all three evaluation seeds. Almost all of the tokens are spent by the predictor scoring candidate libraries on the dev split, and the optimizer accounts for under $0.5\%$. This is far less than validating the same candidates with live episodes. On L1, running each of the 32 candidates on the 138 dev tasks would take 4,416 live episodes, or about 2.8B input tokens at \method{}'s 636.9K per episode, which is 16 times our total. The same comparison gives about 75 times on L2 and 5 times on Mind2Web. Live validation would also need a resettable environment for every dev task, which logs do not provide. We instead score fixed logs and never execute actions, so optimization does not require costly environment setup or iteration.

\noindent\textbf{\method{} trades a longer prompt for higher success.}
\method{} achieves the highest success rate in every column of \cref{tab:main-results}, and this comes with more input tokens. On L1, it uses 636.9K input tokens per episode against 343.4K to 386.1K for the baselines (\cref{tab:token-counts}). The overhead comes from a longer per-step prompt, not from longer episodes. \method{} uses 20.3K input tokens per L1 step, compared with 11.2K to 12.9K for the other methods, while its episode length (31.4 steps) stays within the baseline range of 27.4 to 34.4 steps (\cref{tab:step-counts}). On L2, it takes the fewest steps of any method (90.0 versus 94.3 to 121.2), so it uses about as many input tokens as the naive agent (1.52M versus 1.49M).
The main cause is the library, which is included in the prompt at every step. The final L1 library has 56 skills and 6.4K tokens, and re-sending it on each of about 31 steps accounts for roughly 201K tokens per episode. We also tried giving the Random-$k$ baseline more input tokens to see whether that closes the gap. With $k{=}30$, Random-$k$ uses 504.2K input tokens per L1 episode but reaches only $63.7\%$, below its $65.1\%$ with $k{=}8$.
We also note that while output tokens are usually more expensive than input, the output length is almost unaffected (3.1K tokens per L1 episode, within the 2.3K to 3.4K range of the baselines). Since the library is a fixed part of the prompt, prompt caching can reuse it across steps, and a per-task skill retriever, analogous to workflow-based retrieval, could reduce the per-step cost further. We leave this to future work.
\section{Related Work}
\noindent\textbf{Learning Agent Skills From Experience.}
A growing body of work improves LLM agents with reusable knowledge learned from experience, including executable skills and abstractions \citep{wang2023voyager,stengeleskin2024regal,zheng2025skillweaver}, natural-language insights and trajectory-derived guidance \citep{shinn2023reflexion,zhao2024expel,ouyang2025reasoningbank,ni2026trace2skill}, retrieved past trajectories \citep{zheng2024synapse}, and hindsight rewrites of failed trajectories \citep{hu2026sampleefficientonlinelearninglm}.
Other work uses reinforcement learning to improve skill use and generation \citep{wu2025evolver,xia2026skillrl,wang2026sage,ouyang2026skillos}, or refines prompts, skills, and other artifacts with execution feedback \citep{agrawal2025gepa,alzubi2026evoskill,zhang2026coevoskills,qiu2026autorefine}. Like these methods, our optimization is a gradient-free, discrete search over edits, but it edits a structured library and scores candidates by teacher-forced action prediction on fixed logs. Closest to our procedure, SkillOpt \citep{yang2026skillopt} keeps bounded edits to a skill document that improve held-out task performance, but it validates each candidate by executing tasks. \method{} instead uses skill-guided progress, which needs no live environment or new execution outcomes.

\noindent\textbf{Learning From Demonstrations and Activity Logs.}
Learning procedural knowledge from recorded behavior connects to imitation learning \citep{ross2011dagger}, goal-conditioned learning \citep{andrychowicz2017hindsight}, and temporal abstraction through options \citep{klissarov2021flexible}. Closer to our setting, AWM \citep{wang2024awm} induces textual workflows from goal-labeled trajectories, and NNetNav \citep{murty2024nnetnav} explores websites without predefined goals, labels the resulting trajectories with instructions, and fine-tunes a browser agent. Robotic process mining discovers recurring routines in user-interface logs, including unsegmented logs with unrelated events \citep{leno2021rpm,leno2020unsegmented}. SkillClaw \citep{ma2026skillclaw} learns a shared skill repository from multi-user agent interactions, but its logs include task goals, and its validation reruns tasks in users' environments. Our work differs from these approaches in two ways. First, we learn a textual skill library from logs without recorded task goals. Second, we evaluate updates against recorded actions without task replay, even when the logs interleave several tasks.
\section{Conclusion}
We introduced \method{}, a framework that learns a textual skill library from offline logs that record no goals, cannot be replayed, and may interleave tasks. Using the recorded actions as supervision, \method{} infers a goal for each trajectory, proposes library edits from action-prediction errors, and keeps an edit only when it raises skill-guided progress on held-out trajectories. At test time, the learned workflows retrieve demonstrations that cover a new task's subgoals. \method{} outperforms retrieval and workflow-induction baselines on WorkArena and Online-Mind2Web, raising average success over the strongest baseline by $6.7\%$ and $7.7\%$. It also leads with Qwen-3.6 and matches the strongest baseline with Mistral-3.5. On WorkArena L1, the live success rate rises with skill-guided progress during optimization ($74.1\% \rightarrow 92.3\%$). \method{} keeps the highest average success under the heaviest task interleaving. Skill optimization and workflow-based retrieval are complementary, and scoring edits on fixed logs costs 5 to 75 times fewer tokens than live validation. Overall, these results suggest that the logs people already leave while using software can teach an agent reusable skills and show whether a skill update helps without a live environment.

\bibliographystyle{abbrvnat}
\bibliography{references}

\clearpage
\appendix
\crefalias{section}{appendix}
\crefalias{subsection}{appendix}
\crefalias{subsubsection}{appendix}

\section*{Appendix}
\section{Prompts}
\label{app:prompts}
The goal-inference prompt (\cref{app:goal-inference}) gives each training trajectory the inferred goal $\hat{g}$ that the predictor conditions on (\cref{sec:problem}), and the workflow-induction prompt (\cref{app:workflow-induction}) builds the initial workflows of the library (\cref{sec:skilltune}). During optimization, the optimizer proposes edits with the prompt in \cref{app:optimizer-prompt}, and the predictor predicts actions with the prompt in \cref{app:predictor-prompt}. At test time, the agent receives the help text in \cref{app:agent-prompt}, and workflow-based retrieval uses the prompts in \cref{app:retrieval-prompts}. Placeholders in angle brackets, such as \texttt{<SKILL\_LIBRARY>}, stand for content filled in at run time.

\subsection{Goal-Inference Prompt}
\label{app:goal-inference}
Recall that the training trajectories carry no goals, so during skill optimization the predictor receives an inferred goal $\hat{g}$ (\cref{sec:problem}). Before optimization, we infer one goal per train and dev trajectory with the backbone at temperature 0. WorkArena and Mind2Web use the two system prompts below, and both share the user message that follows. If an output breaks a rule, for example by exceeding the word limit or listing UI actions, we append the reason to the user message and ask again, up to five attempts. The inferred goals are cached and reused across optimization steps.

\begin{example*}[Goal-Inference Prompt for WorkArena]
\begin{lstlisting}[style=skilldump]

You reconstruct the most specific plausible user request behind one successful ServiceNow/WorkArena UI trajectory instance. Produce a concise desired-outcome sentence, not instructions for carrying it out.

Grounding and anti-leakage rules:
- Output exactly <intent>ONE sentence of at most 50 words</intent> and nothing else. The tags are formatting and are removed before the intent is stored.
- Preserve every task-defining detail recoverable from the trajectory: the target record or object type (incident, problem, change request, hardware asset, user, expense line, catalog item, ...), the operation performed, and the concrete values that shaped it -- short descriptions, categories, impact, urgency, priority, filter criteria, amounts, dates, or quantities. Exact task values are useful and should not be replaced by vague words such as "specified", "chosen", or "desired".
- Infer only details supported by the actions; a filled field or selected option expresses a user constraint, not merely a UI instruction.
- Do not list, paraphrase, or imply an ordered sequence of UI actions.
- Do not use action verbs as UI commands (click, fill, select, press, navigate).
- Do not mention selectors, element IDs, application or module names, generic control labels, or the mechanics of operating the page.
- Do not expose passwords, personal contact details, authentication data, or opaque record numbers; describe those only as the required account or contact information.
- If a detail is uncertain, omit only that detail; do not generalize away other constraints the trace clearly supports.

Good: <intent>Create a hardware incident for an overheating laptop with high impact and medium urgency.</intent>
Bad (too vague): <intent>Create an incident record with the specified details.</intent>
Bad (action recipe): <intent>Open Incident, fill Short description, then set Impact and Urgency.</intent>
\end{lstlisting}
\end{example*}

\begin{example*}[Goal-Inference Prompt for Mind2Web]
\begin{lstlisting}[style=skilldump]

You reconstruct the most specific plausible user goal behind one successful Mind2Web trajectory. Produce a concise desired-outcome sentence, not instructions for carrying it out.

Grounding and anti-leakage rules:
- Output exactly <intent>ONE sentence of at most 50 words</intent> and nothing else. The tags are formatting and are removed before the intent is stored.
- Preserve every non-sensitive task-defining detail recoverable from the trajectory: named products, events, venues and destinations; dates or time ranges; quantities and variants; categories, filters, accessibility needs, sort preferences, and requested final state. Exact task values are useful and should not be replaced by vague words such as "specified", "chosen", or "desired".
- Infer only details supported by the actions. A clicked event, category, filter, product, or result may express a user constraint rather than merely a UI instruction.
- Do not list or imply an ordered sequence of UI actions.
- Do not use action verbs as UI commands (click, fill, select, press, navigate).
- Do not mention selectors, element IDs, generic control labels, or the mechanics of operating the page.
- Do not expose passwords, payment-card data, personal contact details, authentication data, or opaque record identifiers. Describe those only as required contact, payment, or account information.
- If a detail is uncertain, omit only that detail; do not generalize away other constraints that the trace clearly supports.

Good: <intent>Find the cheapest wheelchair-accessible parking near Busch Stadium for the Pittsburgh Pirates at St. Louis Cardinals event.</intent>
Bad (too vague): <intent>Find and filter accessible parking options near a specified venue.</intent>
Bad (action recipe): <intent>Fill the parking search, click the event, open Filter, choose wheelchair, and sort by price.</intent>
\end{lstlisting}
\end{example*}

\subsection{Workflow-Induction Prompt}
\label{app:workflow-induction}
Recall that the library starts with workflows only, before any skill is added (\cref{sec:skilltune}). We initialize the workflows $\mathcal{W}$ with a single call to the backbone over all training trajectories. The backbone sees only the recorded action sequences, because the trajectories carry no goals. WorkArena and Mind2Web use the two prompts below. Each prompt consists of an instruction and a one-shot example, and the recorded action sequences follow in the format of the last box. The original Mind2Web one-shot example lists a goal above each action sequence, and we remove those lines so that the example matches the goal-free input.

\begin{example*}[Workflow Induction Prompt for WorkArena]
\begin{lstlisting}[style=skilldump]

### Role & objective
You are a workflow manager for a WorkArena agent operating in ServiceNow. You are given a batch of action sequences that completed enterprise tasks in that environment. Your job is to distill this batch into a library of reusable sub-routines ("workflows") that a future agent can compose to solve unseen tasks.

### What makes a good workflow
A workflow captures a recurring intent (e.g. "create_a_record", "apply_a_filter", "sort_a_list"), not a single task instance. Extract a workflow only when the same action pattern appears, i.e., allow for parameter differences across two or more distinct tasks.

For each workflow:
1. Include at least two ordered steps.
2. Replace anything that varies between instances with a typed placeholder in angle brackets: <LIST_NAME>, <FIELD>, <VALUE>, <OPERATOR>, <DIRECTION>.

After drafting, verify: (1) every workflow is grounded in two or more distinct input action sequences. (2) no two workflows overlap in purpose, (3) placeholders cover only variable elements, (4) you have generate as many workflows as needed to cover all the tasks. Silently revise until all four hold.

### Output format
Return only the workflow library, nothing else:

Workflow: <name>
Description: <when to use this workflow: the goal/intent it serves, any preconditions or entry state it assumes, and the outcome it produces>
1. <action>          # comment with example value(s)
2. <action>
...

Workflow: sort_list
Description: Use when a ServiceNow list must be ordered by one or more fields. Preconditions: the target list is visible. Open the list builder, add one sort row per requested field, and set each row's direction. Outcome: the visible records are ordered by the configured sort levels.
1. click('.list_filter_toggle')
2. click('role=button')  # add a sort row
3. select_option(<SORT_ROW>, <FIELD>)  # e.g. 'nth=1', 'Assigned'
4. select_option(<SORT_ROW>, <DIRECTION>)  # e.g. 'nth=1', 'ascending'
5. Repeat steps 2-4 for each additional sort level  # e.g. 'City', 'descending'
6. click('Run filter')
\end{lstlisting}
\end{example*}

\begin{example*}[Workflow Induction Prompt for Mind2Web]
\begin{lstlisting}[style=skilldump]

### Role & objective
You are a workflow manager for a Mind2Web agent operating across many unrelated public websites. You are given offline Mind2Web gold action sequences expressed as click(<VISIBLE_LABEL>), fill(<VISIBLE_LABEL>, <VALUE>), and select_option(<VISIBLE_LABEL>, <OPTION>). Your job is to distill this batch into a library of reusable, website-agnostic sub-routines ("workflows") that a future agent can compose to solve unseen tasks on unfamiliar sites.

### What makes a good workflow
A workflow captures a recurring cross-site intent (e.g. "complete_contact_form", "search_then_filter", "select_item_and_open_details"), not a single website or task instance. Extract a workflow only when the same semantic action pattern appears across two or more distinct tasks. Generalize site-specific visible labels when they vary; do not invent controls or selectors absent from the input.

For each workflow:
1. Include at least two ordered steps.
2. Replace anything that varies between instances with a typed placeholder in angle brackets: <FIELD>, <VALUE>, <OPTION>, <ITEM>, <SUBMIT_CONTROL>. Keep a visible control label literal only when it is invariant in the supporting traces; otherwise parameterize it. Empty or noisy labels in a trace are not reusable selectors.

After drafting, verify: (1) every workflow is grounded in two or more distinct input action sequences, (2) no two workflows overlap in purpose, (3) placeholders cover only variable elements, (4) you have generated as many workflows as needed to cover all the tasks. Silently revise until all four hold.

### Output format
Return only the workflow library, nothing else:

Workflow: <name>
Description: <when to use this workflow: the goal/intent it serves, any preconditions or entry state it assumes, and the outcome it produces>
1. <action>          # comment with example value(s)
2. <action>
...

## Concrete Examples
Actions:
  click("")
  fill("First Name (required)", "Allan")
  fill("Last Name (required)", "Smith")
  fill("Email Address (required)", "allan.smith@gmail.com")
  fill("Confirm Email Address (required)", "allan.smith@gmail.com")
  fill("ZIP Code (required)", "10001")
  click("Submit")

Actions:
  click("Demo Drive")
  click("Model Y")
  fill("Last Name", "Adams")
  fill("First Name", "Roy")
  fill("Phone Number", "123-999-0000")
  fill("Email Address", "RA@gmail.com")
  fill("Zip Code", "90001")
  click("Submit and Continue")

## Summary Workflows
Workflow: complete_contact_form
Description: Use when a public website asks for a person's contact details before a signup or request can be submitted. Preconditions: the contact form is visible. Outcome: the supplied details are submitted and the site advances to confirmation or the next step.
1. fill(<FIRST_NAME_FIELD>, <FIRST_NAME>)  # e.g. "First Name (required)"
2. fill(<LAST_NAME_FIELD>, <LAST_NAME>)  # e.g. "Last Name"
3. fill(<EMAIL_FIELD>, <EMAIL>)  # e.g. "Email Address"
4. fill(<ADDITIONAL_CONTACT_FIELD>, <VALUE>)  # repeat only for requested visible fields
5. click(<SUBMIT_CONTROL>)  # e.g. "Submit", "Submit and Continue"
\end{lstlisting}
\end{example*}

\subsection{Optimizer Prompt}
\label{app:optimizer-prompt}
At every optimization step, the optimizer receives the system prompt below, followed by a stage-specific suffix, and the user message in the last box of this subsection. The prompt is shared across stages and still describes a trace buffer, which is disabled in all our runs, so the optimizer never sees a trace menu. The suffix restricts the allowed operations to the current stage. \texttt{<MAX\_EDITS>} is the per-step edit limit (\cref{tab:hyperparameters}). On Mind2Web, the first sentence ends with ``general web-navigation tasks across many unrelated public websites'' instead of ``ServiceNow / WorkArena tasks'', and the Mind2Web rules box is appended to the system prompt. On WorkArena, the page evidence of each case reads ``(unavailable)'', because the recorded trajectories hold no page observations. A \texttt{distill\_workflow} or \texttt{replace\_workflow} edit is resolved by a second call with the workflow-distillation prompt at the end of this subsection.

\begin{example*}[Optimizer Prompt for WorkArena]
\begin{lstlisting}[style=skilldump]

You improve a reusable SKILL that helps a web agent predict the correct next action on ServiceNow / WorkArena tasks. The skill has THREE editable sections and you are given a batch of cases the agent got WRONG: for each, the goal, the actions already taken, the agent's PREDICTED next action, and the CORRECT next action.

Editable sections:
1. SKILLS -- a numbered list of general guidance. Edit with add / replace / delete (ids refer to the skill numbers).
2. WORKFLOWS -- a library of reusable parameterised sub-routines induced from real traces. You do NOT hand-write these; instead:
   - distill_workflow: name a task template and a fresh workflow is induced from that task's oracle traces and appended.
   - replace_workflow: replace workflow #id with one freshly distilled from the named task.
   - delete_workflow: remove workflow #id.
3. TRACES -- a buffer of reference trajectories injected as in-context examples. Starts EMPTY and grows as you see failures: add_trace pulls in a pool menu item, delete_trace evicts a slot that is no longer useful, and replace_trace swaps one slot for a better menu item in a single move. The buffer holds at most 8 traces at a time -- once full you must delete_trace or replace_trace before adding more.

Use the RIGHT lever for each error pattern:
- A wrong value / which-field / verb choice that GENERALISES -> a SKILLS edit.
- Every added or replaced SKILL must preserve the general rule and then include
    one concrete supporting example copied from the mispredicted cases, using
    this exact shape: "<general rule> Example: for goal '<goal>', after
    <relevant actions or 'no prior actions'>, the correct next action is
    <gold action> (not <predicted action>)." The example is evidence from a
    TRAINING reflection case, not a claim about the untouched test set. Quote the
    actual goal and actions closely enough that a reader can audit which mistake
    motivated the skill; concrete values are allowed in the Example clause, but
    the general rule before it must remain reusable.
- A missing procedure for a whole task family -> distill_workflow for that task.
- Workflows must be compositional and uniquely named: prefer reusable UI operations over table-specific or fixed-cardinality variants, and replace or delete an overlapping workflow instead of adding another name for it.
- The in-context examples don't cover the failing task family -> add_trace (or replace_trace if the buffer is full) to bring in a relevant example (pick a menu item whose task matches the failing cases).
- An existing trace is redundant or irrelevant to any recent failure -> delete_trace to make room.

Hard constraints:
- Do NOT hard-code instance-specific values (names, numbers, ids) in the
general-rule part of a skill; those do not generalize. They may appear only in
the required concrete Example clause.
- Keep each skill concise, imperative, and <= 600 characters including its
required Example clause.
- Prefer editing/merging existing items over piling on near-duplicates.
- Propose at most <MAX_EDITS> edits.

Reply with STRICT JSON only, no prose, no code fences:
{"edits": [
  {"op": "add", "text": "<new skill>", "reason": "<why>"},
  {"op": "replace", "id": <skill#>, "text": "<skill>", "reason": "<why>"},
  {"op": "delete", "id": <skill#>, "reason": "<why>"},
  {"op": "distill_workflow", "task": "<task_name>", "reason": "<why>"},
  {"op": "replace_workflow", "id": <workflow#>, "task": "<task_name>", "reason": "<why>"},
  {"op": "delete_workflow", "id": <workflow#>, "reason": "<why>"},
  {"op": "add_trace", "add": <menu_id>, "reason": "<why>"},
  {"op": "replace_trace", "evict": <slot#>, "add": <menu_id>, "reason": "<why>"},
  {"op": "delete_trace", "evict": <slot#>, "reason": "<why>"}
]}
If no edit would help, reply {"edits": []}.
\end{lstlisting}
\end{example*}

\begin{example*}[Optimizer Prompt for Mind2Web]
\begin{lstlisting}[style=skilldump]

Mind2Web-specific rules:
- A case header has the form "Case N [task_name]". Its task_name is one of the
    train-derived semantic intent clusters (for example travel_search or
    contact_form_completion) and is the exact key required by the trace pool.
- When a goal is blank, first infer internally what the task is about from the
    action history and CURRENT PAGE EVIDENCE. Use action verbs, visible labels,
    values already present in the history, and ordering to identify reusable
    guidance. The correct next action is a training label used to diagnose the
    error, not information the deployed predictor will possess. Never write a
    general rule that requires knowing a requested entity, value, destination,
    or goal that is absent from both the displayed goal and prior actions. In
    particular, with a blank goal and no prior actions, do not invent hidden
    task intent from the correct action or task_name; propose only a genuinely
    page-observable default, or no skill edit.
- For SKILLS, ground the general rule in distinctions visible in CURRENT PAGE
    EVIDENCE and the action history. The semantic task_name and correct action
    are optimizer-only TRAIN metadata and must not become hidden prerequisites
    of the rule. Treat task_name as a broad family, not as a full description
    of the current page or instance.
- For distill_workflow or replace_workflow, copy `task` EXACTLY from one of the
    supplied case-header task names. Do not invent a narrower family name;
    describe refinements in `reason` instead. This exact-copy rule lets the
    resolver retrieve all TRAIN oracle traces in that semantic cluster.
\end{lstlisting}
\end{example*}

\begin{example*}[Optimizer Suffix for the Workflow Stage]
\begin{lstlisting}[style=skilldump]

THIS ROUND you may use ONLY these ops: delete_workflow, distill_workflow, replace_workflow. Propose no other op type; reply {"edits": []} if none of these would help.

WORKFLOW-ONLY ROUND: every supplied case is evidence that the current workflow library failed to guide the correct next action. For this round, OVERRIDE the earlier right-lever rule that assigns wrong field, value, or verb choices to SKILLS; skills are unavailable, so express a plausible procedural correction by replacing the matching workflow. Existing coverage of a case's task family is NOT by itself a reason to return no edits. If a matching workflow exists, propose replace_workflow using that workflow's id and copy the case-header task_name exactly. Use distill_workflow only when no existing workflow covers the family, and delete_workflow only for clear redundancy. You propose candidates; a downstream evaluation gate, not you, decides whether they help. With non-empty cases, do not return an empty edit list unless no legal workflow op can be formed from the displayed workflows and case task names.
\end{lstlisting}
\end{example*}

\begin{example*}[Optimizer Suffix for the Skill Stage]
\begin{lstlisting}[style=skilldump]

THIS ROUND you may use ONLY these ops: add, delete, replace. Propose no other op type; reply {"edits": []} if none of these would help.

ID SCOPE: 'id' in replace/delete refers ONLY to the SKILLS numbering (### Skills). The ### Workflows list is READ-ONLY context this round -- never use a workflow number as a skill id. Do not propose delete/replace for a skill id that is not present in the current SKILLS list.
PREFER ADDING new, DISTINCT skills that cover error patterns not yet addressed. Do NOT repeatedly rewrite an existing skill that is already working -- only replace one if it is clearly wrong or superseded. Every add/replace text MUST contain an 'Example:' copied from one of the supplied training reflection cases, including that case's goal and correct next action.

SKILLS-OVER-WORKFLOWS ROUND: the displayed workflows are already active READ-ONLY guidance. Add a skill only for a decision, disambiguation, or recovery rule that the workflows do not encode. Do not paraphrase a workflow's procedure into a broad skill; duplicated guidance adds prompt noise and can make action selection worse.
\end{lstlisting}
\end{example*}

\begin{example*}[Workflow Distillation Prompt]
\begin{lstlisting}[style=skilldump]

You are a workflow manager for a web-navigation agent. From the successful action traces of a single ServiceNow / WorkArena task template, distill ONE reusable sub-routine ("workflow") that a future agent can compose to solve unseen tasks.

A workflow captures a recurring intent (e.g. "create_a_record", "apply_a_ filter", "sort_a_list"), not one task instance. Capture the COMMON action sequence shared across the given traces; drop steps that appear in only one trace.

Requirements:
- Give it a verb-based snake_case name describing the operation, never a table, record, seed, or count (apply_filter, not filter_incidents_by_priority).
- Include at least two ordered steps in the action vocabulary (click/fill/select_option/set_checked/...).
- Replace anything that varies between instances with a typed placeholder in angle brackets: <LIST_NAME>, <FIELD>, <VALUE>, <OPERATOR>, <DIRECTION>. Keep invariant labels (menu/list/button names such as "All", "New", "Submit") verbatim.
- Make repeated structure compositional: say "for each condition/field/sort key" and parameterise its field, operator, connector, value, and direction. Never hard-code the number of conditions, fields, keys, or repeated steps.
- Prefer the shortest correct sequence; do not invent steps not supported by the traces.

Output ONLY the single workflow, nothing else (no prose, no code fences):

Workflow: <name>
Description: <one line: when to use this>
1. <action>          # comment with example value(s)
2. <action>
...
\end{lstlisting}
\end{example*}

\subsection{Predictor Prompt}
\label{app:predictor-prompt}
The predictor uses a benchmark-specific system prompt. When the library is non-empty, we append the header \texttt{\#\# Skill (learned guidance for this task family)} and then the library, i.e., the numbered skills (in the skill stage) followed by the workflows. The user message holds the inferred goal and the history. On Mind2Web, it also starts with the recorded page observation, i.e., the website, URL, pruned accessibility tree, candidate elements, cleaned DOM, and visible text. The WorkArena user message has no observation block.

\begin{example*}[Predictor Prompt for WorkArena]
\begin{lstlisting}[style=skilldump]

You are simulating a ServiceNow / WorkArena web agent. Given the task goal and the browser actions already taken, output the SINGLE next action that makes progress toward completing the goal.

Action vocabulary -- use the element's VISIBLE LABEL as the first argument (never a numeric id):
  click("<label>")
  fill("<label>", "<value>")
  select_option("<label>", "<value>")
  set_checked("<label>", "<true|false>")
  press("<label>", "<key>")
  noop(<milliseconds>)

Rules:
- Output ONLY the single next action call, on one line. No prose, no code fences.
- Take field labels and values verbatim from the goal where applicable.
- Predict exactly one action (the immediate next step), not the whole plan.
\end{lstlisting}
\end{example*}

\begin{example*}[Predictor Prompt for Mind2Web]
\begin{lstlisting}[style=skilldump]

You are simulating a general web-navigation agent across many unrelated public websites. Given the task goal (when available) and the browser actions already taken, output the SINGLE next action that makes progress toward the task.

Action vocabulary -- use the element's VISIBLE LABEL as the first argument (never a numeric id, CSS selector, XPath, or backend-node id):
    click("<label>")
    fill("<label>", "<value>")
    select_option("<label>", "<value>")
    set_checked("<label>", "<true|false>")
    press("<label>", "<key>")
    noop(<milliseconds>)

Rules:
- Output ONLY the single next action call, on one line. Do not output your inference, reasoning, prose, or code fences.
- Use visible labels and task-specific values supported by the goal and history.
- Do not assume controls from an enterprise app or transfer labels from an unrelated website.
- Predict exactly one action (the immediate next step), not the whole plan.
- Ground the action in the current Mind2Web page observation. Prefer candidate
visible labels and accessible names; use the pruned tree, nearby cleaned DOM,
URL/website, and visible page text to disambiguate controls. Never output a
candidate id as the action target.
- When the goal is blank, first infer internally what kind of task or reusable workflow is underway from the actions already taken: their verbs, visible labels, entered values, and order. Then use that inferred intent together with the learned workflows and skills to choose the next action. Revise the inferred intent as more actions arrive; do not emit the inferred intent.
- When both the goal and action history are empty, no task-specific intent can be inferred. Follow only broadly applicable learned guidance; never invent a task-specific value or claim that an unseen control is present.
\end{lstlisting}
\end{example*}

\subsection{Agent Prompt}
\label{app:agent-prompt}
At test time, the WorkArena agent is the AgentLab GenericAgent, and the Online-Mind2Web agent is a Browser Use agent. We keep their own system prompts and add the help text below. On WorkArena, the help text goes into the agent's extra instructions, after the task goal that AgentLab provides, and all $k$ retrieved demonstrations appear in one list without their goals. On Online-Mind2Web, the help text is placed before the task, and matched demonstrations are listed with the subgoals they cover, separately from the randomly sampled ones.

\begin{example*}[Agent Prompt for WorkArena]
\begin{lstlisting}[style=skilldump]

You are about to solve a Web Agentic task. You are given two kinds of optional reference material below. (1) Skills and Workflows. These are HIGHER LEVEL: it describes reusable procedures, behavioral guidance, and overall step patterns, and may omit low-level interaction details. (2) The REFERENCE TRAJECTORIES are LOWER LEVEL: they contain concrete action sequences from previously attempted tasks. Take the trajectories only as references for how individual UI interactions can be carried out when applicable. They are not necessarily similar to the current task, are not guaranteed to be relevant, and are not instructions to follow verbatim. Variable placeholders like <RECORD_NAME> or <VALUE> stand for instance-specific values; replace them with values from the current task.

## Workflow memory
<SKILL_LIBRARY>

## Reference trajectories
--- Reference trajectory 1 ---
Actions (<N> steps):
  1. <ACTION_1>
  ...

--- Reference trajectory 2 ---
...

End of help. Now act on the current task; do not mention this help in your output.
\end{lstlisting}
\end{example*}

\begin{example*}[Agent Prompt for Online-Mind2Web]
\begin{lstlisting}[style=skilldump]

Below is a web-navigation skill learned entirely from offline Mind2Web training trajectories. Apply only the relevant workflows and rules; adapt labels and values to the current page.

<SKILL_LIBRARY>

End of learned skill. Now choose the next action for the current task.

Subgoal coverage:
Reference 1 supports: <SUBGOAL>; <SUBGOAL>
...

Below are task-matched reference trajectories selected because they cover different reusable subgoals of the current task. Their goals are withheld. Compose only the relevant parts; do not treat any complete trajectory as a recipe for the live site.

--- Reference trajectory 1 ---
Actions (<N> steps):
  1. <ACTION_1>
  ...
End of task-matched workflow references.

Below are additional exploration trajectories sampled at RANDOM. They are not claimed to match the current task. Use them only as generic examples of available UI actions when the task-matched patterns are incomplete.

--- Reference trajectory 1 ---
...
End of random exploration references.

Use the reference only as hints; the live site may differ.

## Your task
<TASK_GOAL>
\end{lstlisting}
\end{example*}

\subsection{Retrieval Prompts}
\label{app:retrieval-prompts}
Workflow-based retrieval (\cref{app:retrieval}) uses two prompts, both at temperature 0. Offline, the annotation prompt labels each training trajectory with the workflows it instantiates, seeing only the trajectory's actions. For each test task, the planner prompt decomposes the task goal into subgoals and maps each to workflows. The workflow inventory holds the library's workflows, except that the WorkArena planner receives the whole library, including its skills.

\begin{example*}[Workflow Annotation Prompt]
\begin{lstlisting}[style=skilldump]

You label which reusable WORKFLOWS a web-navigation solution uses.

You are given an inventory of named workflows (each a "Workflow: <name>" header
with numbered steps) and ONE item to label -- either an action trace or a task
goal. Return the subset of workflow NAMES that the item instantiates / would
require.

Rules:
- Output ONLY workflow names from the inventory, one per line, verbatim (the
    snake_case name after "Workflow:"). No numbering, prose, or code fences.
- If no inventory workflow applies, output the single sentinel NONE. Never
    return an empty response.
- Include a workflow only if the item clearly uses (trace) or requires (goal)
    it. Omit ones that do not apply. It is fine to return several names.
- Do not invent names that are absent from the inventory.
\end{lstlisting}
\end{example*}

\begin{example*}[Planner Prompt]
\begin{lstlisting}[style=skilldump]

Decompose one web-navigation goal into the ordered reusable workflows needed to
complete it.

Return ONLY a JSON object with this shape:
{"subgoals":[{"text":"short action-oriented subgoal","workflows":["name"]}]}

Rules:
- Preserve execution order and every explicit constraint (search, dates,
  filters, sorting, selection, cart/checkout, and so on).
- Use only workflow names from the supplied inventory, verbatim.
- A subgoal may have several workflows or an empty list when the inventory has
  no adequate workflow. Never invent a workflow name.
- Prefer 1-6 concise subgoals. Do not include concrete solutions or UI labels.
- Do not wrap the JSON in markdown.
\end{lstlisting}
\end{example*}

\section{Workflow-Based Retrieval Details}
\label{app:retrieval}
Recall that in \cref{sec:workflow-retrieval}, we select demonstrations that together cover the subgoals of a new task, using the optimized workflows $\mathcal{W}^{*}$ as an intermediate representation. We describe the full procedure here in four steps: (1) annotating training trajectories with workflows offline, (2) planning subgoals for a new task, (3) scoring each trajectory against each subgoal, and (4) greedily selecting trajectories that cover the subgoals.

\noindent\textbf{Workflow Annotation.}
Offline, we annotate each training trajectory $\tau$ with the workflows it instantiates:
\begin{equation}
    Z(\tau)
    \subseteq
    \mathcal{W}^{*}.
\end{equation}
For example, a trajectory that searches for an item, applies a filter, and submits a form may be annotated with three corresponding workflows. Each workflow $w\in\mathcal{W}^{*}$ then carries a weight that decreases with the number of training trajectories that instantiate it. A workflow shared by most of the training trajectories therefore contributes little to a match, while a rare one contributes most. The weight enters the workflow component of the relevance score below.

\noindent\textbf{Subgoal Planning.}
For a new task, a planner decomposes the task goal into an ordered collection of subgoals:
\begin{equation}
    P(g)
    =
    \left[
        (u_1,W_1),
        \ldots,
        (u_m,W_m)
    \right],
\end{equation}
where $u_j$ is a subgoal and $W_j\subseteq\mathcal{W}^{*}$ contains workflows that may accomplish it. We therefore compare training trajectories and task goals in a shared procedural space.

\noindent\textbf{Relevance Scoring.}
We assign each trajectory and subgoal pair a relevance score
\begin{equation}
    R(\tau,u)
    =
    \operatorname{Combine}
    \left(
        R_{\mathrm{workflow}},
        R_{\mathrm{semantic}},
        R_{\mathrm{site}}
    \right).
\end{equation}
The workflow component measures weighted overlap between $Z(\tau)$ and the workflows associated with $u$. The semantic component separates trajectories that instantiate the same broad workflow for different purposes. The site component adds a small preference for demonstrations from a compatible website. We treat only sufficiently confident matches as task-relevant.

\noindent\textbf{Greedy Selection.}
Let $U$ be the set of subgoals that remain uncovered. The selector repeatedly chooses the trajectory that best covers the current set:
\begin{equation}
    \tau^{*}
    =
    \operatorname*{arg\,max}_{\tau}
    \operatorname{Coverage}(\tau,U).
\end{equation}
After selecting $\tau^{*}$, we remove the subgoals it supports from $U$. We stop when the task is covered, no confident match remains, or the demonstration budget ($k{=}8$, \cref{sec:setup}) is reached.

The matched set $\mathcal{E}_{\mathrm{matched}}(g)$ contains demonstrations supported by workflow evidence. The exploration set $\mathcal{E}_{\mathrm{explore}}(g)$ contains a small number of deterministically sampled examples, and retrieval slots not filled by confident matches are assigned to exploration. If no workflow can be associated with the task, then $\mathcal{E}_{\mathrm{matched}}(g)=\varnothing$ and we fall back entirely to deterministic random retrieval. On Mind2Web, the prompt labels each matched example with the subgoals it supports and presents exploratory examples separately as generic demonstrations. On WorkArena, all $k$ demonstrations appear in one list without subgoal labels (\cref{app:agent-prompt}).

\section{Hyperparameters}
\label{app:hyperparameters}
\cref{tab:hyperparameters} lists the settings of skill optimization, workflow-based retrieval, and the test-time agent. All runs use inferred goals and the two-stage schedule of \cref{sec:skilltune}, and they validate on the full dev split without a curriculum. Unless noted, the settings are shared across WorkArena L1, WorkArena L2, and Mind2Web and across backbones. The Mistral-3.5 deployment samples every call, including the predictor and the agent, at temperature 0.7 and top-$p$ 0.95.

\begin{table}[t]
\centering
\caption{Hyperparameters of \method{}. Values are shared across settings and backbones.}
\label{tab:hyperparameters}
\small
\renewcommand{\arraystretch}{1.15}
\begin{tabular}{p{0.36\linewidth} p{0.56\linewidth}}
\toprule
\textbf{Setting} & \textbf{Value} \\
\midrule
\multicolumn{2}{l}{\emph{Skill optimization}} \\
Stage stopping rule & 5 consecutive steps without an accepted edit, or 25 steps \\
Maximum total steps & 100 \\
Failure search & batch size is 12 until at least 6 fail, at most 48 per step \\
Failure cases per step & At most 6, lowest $Q$ first; each shows the last 8 actions \\
Edits per step & At most 3 \\
Library limits & 100 skills, 40 workflows, 600 characters per skill; \\
Acceptance rule & Mean $Q$ on the full dev split strictly increases \\
Predictor sampling & Temperature 0, at most 2,048 output tokens \\
Optimizer sampling & Temperature 0.7, at most 2,048 output tokens \\
\midrule
\multicolumn{2}{l}{\emph{Workflow-based retrieval}} \\
Demonstrations & $k=8$ \\
Planner sampling & Temperature 0, at most 4,096 output tokens \\
Relevance score & $0.65\,R_{\mathrm{workflow}} + 0.25\,R_{\mathrm{semantic}} + 0.10\,R_{\mathrm{site}}$ \\
Workflow weight & IDF over training trajectories, $\ln\frac{1+N}{1+\mathrm{df}}+1$ \\
Confidence threshold & 0.8 \\
\midrule
\multicolumn{2}{l}{\emph{Test-time agent}} \\
Step budget & 100 on WorkArena L1, 200 on others \\
Observation & Text only, no screenshots \\
\bottomrule
\end{tabular}
\end{table}

\section{Data Processing and Statistics}
\label{app:data}
Recall that we learn from successful oracle trajectories on WorkArena and from annotator-recorded trajectories on Mind2Web (\cref{sec:setup}). \cref{tab:data-stats} lists the resulting counts. \cref{sec:setup} describes which WorkArena task families we remove. On both WorkArena levels, the test instances are other seeds of the same task templates as the training instances. We split the WorkArena training trajectories into train and dev by trajectory hash, which gives a roughly even split.
 The Mind2Web training trajectories need more processing, which we describe below.

\noindent\textbf{Mind2Web.}
Online-Mind2Web provides 300 live tasks, each with a start URL and a difficulty label, but no demonstrations. We therefore draw training trajectories from the original Mind2Web training set, which holds 1,009 trajectories. Online-Mind2Web also does not publish the Mind2Web domain and subdomain labels. We label each online task with one Mind2Web category (e.g., Travel $\rightarrow$ Airlines or Shopping $\rightarrow$ Auto) using GPT-5.4 on its instruction and website, and allow a none label when no category fits (13 tasks). We keep a test task only when its category also occurs in the original Mind2Web training set. That set has no Info or Service tasks, so this filter removes all online tasks in those two domains and leaves 122 test tasks (29 easy, 54 medium, and 39 hard) across 17 categories. Applying the same filter in the other direction keeps 966 of the 1,009 training trajectories (7,347 of 7,775 actions), since only Travel $\rightarrow$ Car rental has no matching test task.

We then downsample the 966 trajectories to 640 with a fixed seed so that the category distribution of the training trajectories follows that of the test tasks. An exact match would require duplicate trajectories, because Shopping $\rightarrow$ Digital covers $18.9\%$ of the test tasks but has only 28 training trajectories. We instead cap each category at the trajectories it has and spread the remaining share proportionally over the other categories, so no trajectory appears twice. This lowers the total-variation distance between the training and test category distributions from $0.299$ to $0.204$. The resulting training set has 640 trajectories, 4,677 actions, and 70 websites, which we split 1:1 into train and dev by trajectory hash (\cref{tab:data-stats}).

\begin{table}[t]
\centering
\caption{Training and evaluation data. Trajectory counts are successful oracle (WorkArena) or recorded (Mind2Web) trajectories; action counts are scoreable actions. Train and dev are the trajectory-level hash split used for reflection and edit selection. WorkArena test instances are other seeds of the training task templates. Mind2Web test tasks are live Online-Mind2Web tasks.}
\label{tab:data-stats}
\small
\renewcommand{\arraystretch}{1.2}
\begin{tabular}{l rrr}
\toprule
& \textbf{WorkArena L1} & \textbf{WorkArena L2} & \textbf{Mind2Web} \\
\midrule
Task instances (train / test)       & 270 / 270    & 216 / 198    & 640 / 122 \\
Training actions                    & 4,573        & 3,561        & 4,677 \\
Train split (traj.\ / actions)      & 132 / 2,405  & 103 / 1,995  & 323 / 2,413 \\
Dev split (traj.\ / actions)        & 138 / 2,168  & 78 / 1,566   & 317 / 2,264 \\
\bottomrule
\end{tabular}
\end{table}

\section{Additional Results}
\label{app:additional-results}
We report two sets of results that support \cref{sec:results}: the success rates behind the noise sweep, and token and step counts per episode.

\noindent\textbf{Noise Sweep.}
Recall that in \cref{sec:analysis}, we rebuild every method from perturbed WorkArena training pools with increasing interruption rates $q$. \cref{tab:noise-contrast} lists the success rates plotted in \cref{fig:q_sweep}.

\begin{table}[t]
\centering
\caption{Success rates (\%) on WorkArena as the interruption rate $q$ grows from $0$ (the clean training pool) to $1.0$. Each method uses the same variant at every interruption rate. The $q{=}0$ columns and the Naive row are three-seed means from \cref{tab:main-results}, and all other entries are single-seed. Naive uses no training pool and is unaffected by construction. \textbf{Bold} marks the best result in each column. }
\label{tab:noise-contrast}
\small
\renewcommand{\arraystretch}{1.25}
\begin{tabular}{l cccc cccc}
\toprule
& \multicolumn{4}{c}{\textbf{WorkArena L1}} & \multicolumn{4}{c}{\textbf{WorkArena L2}} \\
\cmidrule(lr){2-5} \cmidrule(lr){6-9}
\textbf{Method} & $q{=}0$ & $q{=}0.33$ & $q{=}0.66$ & $q{=}1.0$
                & $q{=}0$ & $q{=}0.33$ & $q{=}0.66$ & $q{=}1.0$ \\
\midrule
Naive              & 45.6 & 45.6 & 45.6 & 45.6
                   & 48.0 & 48.0 & 48.0 & 48.0 \\
Random-$k$         & 65.1 & 63.9 & 69.6 & \textbf{75.9}
                   & 54.2 & 61.6 & 47.3 & 48.5 \\
AWM                & 75.7 & 74.1 & 60.9 & 49.3
                   & 51.9 & 52.8 & 50.5 & 50.0 \\
AWM $+$ Random-$k$ & 80.3 & 82.8 & 70.4 & 65.9
                   & 60.4 & 60.6 & 53.7 & 55.1 \\
\method{}          & \textbf{92.3} & \textbf{88.9} & \textbf{83.1} & 74.8
                   & \textbf{61.8} & \textbf{65.8} & \textbf{63.3} & \textbf{62.1} \\
\bottomrule
\end{tabular}
\end{table}

\noindent\textbf{Token and Step Counts.}
\label{app:cost}
Recall that in \cref{sec:analysis}, we attribute \method{}'s extra input tokens to a longer per-step prompt, since its episode lengths stay within the baseline range. \cref{tab:token-counts} gives the average input and output tokens per episode, and \cref{tab:step-counts} gives the average number of steps.

\begin{table}
\centering
\caption{Average tokens per episode, reported as \textbf{input\,/\,output} in thousands (K).
WorkArena values are per level (L1/L2); Online-Mind2Web values are split by task difficulty.}
\label{tab:token-counts}
\small
\renewcommand{\arraystretch}{1.25}
\begin{tabular}{l cc ccc}
\toprule
& \multicolumn{2}{c}{\textbf{WorkArena}} & \multicolumn{3}{c}{\textbf{Online-Mind2Web}} \\
\cmidrule(lr){2-3} \cmidrule(lr){4-6}
\textbf{Method} & L1 & L2 & Easy & Medium & Hard \\
\midrule
Naive
  & 386.1\,/\,3.4 & 1489.0\,/\,16.3
  & 76.6\,/\,3.3 & 166.8\,/\,7.4 & 242.9\,/\,10.6 \\
Random-$k$
  & 369.3\,/\,2.8 & 1404.1\,/\,13.8
  & 90.2\,/\,3.5 & 238.1\,/\,9.3 & 308.2\,/\,12.2 \\
AWM
  & 343.4\,/\,2.5 & 1155.0\,/\,13.0
  & 86.1\,/\,3.5 & 222.6\,/\,9.0 & 314.4\,/\,12.5 \\
AWM $+$ Random-$k$
  & 376.4\,/\,2.3 & 1405.0\,/\,13.6
  & 86.8\,/\,3.3 & 227.2\,/\,8.4 & 328.5\,/\,12.4 \\
\method{}
  & 636.9\,/\,3.1 & 1522.8\,/\,13.0
  & 130.1\,/\,3.9 & 326.5\,/\,10.3 & 405.4\,/\,12.6 \\
\bottomrule
\end{tabular}
\end{table}

\begin{table}[h]
\centering
\caption{Average number of steps per episode. WorkArena values are per level (L1/L2);
Online-Mind2Web values are split by task difficulty.}
\label{tab:step-counts}
\small
\renewcommand{\arraystretch}{1.25}
\begin{tabular}{l cc ccc}
\toprule
& \multicolumn{2}{c}{\textbf{WorkArena}} & \multicolumn{3}{c}{\textbf{Online-Mind2Web}} \\
\cmidrule(lr){2-3} \cmidrule(lr){4-6}
\textbf{Method} & L1 & L2 & Easy & Medium & Hard \\
\midrule
Naive
  & 34.4 & 121.2
  & 6.7 & 13.3 & 17.8 \\
Random-$k$
  & 28.6 & 105.9
  & 6.5 & 14.7 & 19.1 \\
AWM
  & 27.4 & 94.3
  & 6.8 & 14.3 & 19.5 \\
AWM $+$ Random-$k$
  & 30.5 & 94.5
  & 6.4 & 13.7 & 18.8 \\
\method{}
  & 31.4 & 90.0
  & 6.7 & 15.4 & 18.7 \\
\bottomrule
\end{tabular}
\end{table}

\newpage
\section{Qualitative Examples of the Learned Skills}
\label{app:learned-skills}
\begin{example*}[Mistral-3.5 Skill Library Learned on WorkArena L1]
\begin{lstlisting}[style=skilldump]

### Skill

1. After setting all filter conditions, always click 'Run' to apply the filter. Example: for goal 'Find hardware assets with install status Missing and unassigned.', after select_option('nth=1', '='), the correct next action is click('Run') (not select_option('nth=1', 'Unassigned')).
2. For create forms, after clicking a field, fill it before clicking another field. Example: for goal 'Create a hardware asset record for an Apple MacBook Pro 15 with serial number SN-6401df5f-8737-4f37-8146-5b14dbdec668...', after click('Model category'), the correct next action is fill("Model category", "Compute") (not click('Model category')).
3. In create forms, populate fields in the order they appear on the form. Example: for goal 'Create a change request to add a network switch to a cabinet with low impact, an unsuccessful close code, and an assignment to CAB Approva.', after no prior actions, the correct next action is click('Justification') (not click('Short description')).
4. After setting all sort conditions, click 'Run filter' (not 'Run'). Example: for goal 'Sort hardware asset records by Assigned in ascending order and by PO number in ascending order.', after select_option('nth=2', 'ascending'), the correct next action is click('Run filter') (not click('Run')).
5. After completing all filter conditions in a list filter, click 'Run' to execute the filter, even if the last condition used a select_option. Example: for goal 'Find user records with class User, notification disabled, location 2-12-1 Ookayama Meguro-ku Tokyo, name Carla Humble, and VIP status false.', after select_option("nth=4", "false"), the correct next action is click("Run") (not click("nth=0")).
6. For knowledge base searches, begin by filling the 'Search' field with the query in quotes. Example: for goal 'Search for knowledge information about the company's stock ticker symbol.', after no prior actions, the correct next action is fill("Search", "\"company's stock ticker symbol\"") (not click("nth=0")).
7. In create forms, always click the first required field before filling any field. Example: for goal 'Create a change request to upgrade OWA-SD-01 to MS Windows Server 2016 with low risk and medium impact.', after no prior actions, the correct next action is click("Number") (not fill("Justification", "Upgrade OWA-SD-01 to MS Windows Server 2016")).
8. After setting all filter conditions in a list filter, click 'Run' even if the last condition was set via fill/click. Example: for goal 'Find hardware assets with install status in stock, model category Compute, and matching specified CI, asset tag, and assigned_to values.', after fill("nth=4", "Compute"), the correct next action is click("Run") (not select_option("nth=4", "Compute"))
9. In create forms, prioritize populating required fields (e.g., Impact, Risk) before optional fields (e.g., Justification). Example: for goal 'Create a change request with low impact, moderate risk, category Other, and close code Successful with issues.', after no prior actions, the correct next action is select_option("Impact", "3 - Low") (not click('Justification')).
10. In create forms, fill fields with the exact value provided in the goal, not a paraphrased or generated one. Example: for goal 'Create a change request to add a network switch to a cabinet with low impact, an unsuccessful close code, and an assignment to CAB Approva.', after click('Justification'), the correct next action is fill("Justification", "fluozirconic compenser playfellow stirrupless used") (not fill("Justification", "Add a network switch to a cabinet"))
11. For knowledge base searches, always start with a fill in the 'Search' field using the exact query from the goal in quotes. Example: for goal 'Find the type of heating system installed in office #456.', after no prior actions, the correct next action is fill("Search", "\"type of heating system in office #456\"") (not click(".list_filter_toggle")).
12. After setting all OR conditions in a list filter, always click 'Run' to execute the filter. Example: for goal 'Filter change requests where type is Standard or risk is High or change model is empty.', after select_option("nth=2", "="), the correct next action is click("Run") (not select_option("nth=2", "empty")).
13. In create forms, always click the first field before filling it, even if it is a required field. Example: for goal 'Create a hardware asset record for an Apple MacBook Pro 17 with serial number SN-a0d6d2f3-1dd0-48cb-9225-4edf814cc19b, vendor Appl...', after click('Model'), the correct next action is click("Vendor") (not fill("Vendor", "Appl")).

### Workflows

Workflow: create_change_request
Description: Use when creating a new Change Request form record. Preconditions: the Change Request form is open in create mode. Enter any needed text, reference, and choice fields, using lookup confirmation clicks after reference fields when needed. Outcome: a new change request is inserted.
1. click(<FIELD>)  # e.g. 'Short description', 'Number', 'Backout plan'
2. fill(<FIELD>, <VALUE>)  # e.g. 'CHG0000048', 'Upgrade NY RAC to Oracle 12c'
3. If <FIELD> is a reference field, click('nth=0') to accept the lookup suggestion  # e.g. after 'Assignment group', 'Configuration item'
4. For choice fields, select_option(<FIELD>, <VALUE>) instead of fill  # e.g. 'Risk'='Moderate', 'Impact'='3 - Low'
5. Repeat steps 1-4 for each required field on the form
6. click('iframe[name="gsft_main"] >> internal:control=enter-frame >> #sysverb_insert')

Workflow: create_hardware_asset
Description: Use when creating a new hardware asset record. Preconditions: the Hardware Asset form is open in create mode. Populate the common identity fields first, then any optional ownership, financial, lifecycle, and location fields. Outcome: a new hardware asset is inserted.
1. click('Model category')
2. fill('Model category', <MODEL_CATEGORY>)  # e.g. 'Compute'
3. click('nth=0')  # accept suggested model category
4. click('Model')
5. fill('Model', <MODEL>)  # e.g. 'Apple MacBook Air 13'
6. click('nth=0')  # accept suggested model
7. click('Vendor')
8. fill('Vendor', <VENDOR>)  # e.g. 'Appl', 'Lenov'
9. click('nth=0')  # accept suggested vendor
10. click('Serial number')
11. fill('Serial number', <SERIAL_NUMBER>)  # e.g. 'SN-1037183e-...'
12. For each additional field, either fill(<FIELD>, <VALUE>) or select_option(<FIELD>, <VALUE>)  # e.g. 'Asset tag', 'Company', 'Expenditure type'
13. If an additional field is a reference field, click('nth=0') after filling to accept the suggestion  # e.g. 'Location', 'Assigned to | nth=0', 'Company'
14. click('iframe[name="gsft_main"] >> internal:control=enter-frame >> #sysverb_insert')

Workflow: create_incident
Description: Use when creating a new Incident record. Preconditions: the Incident form is open in create mode. Populate the required caller and summary first, then add optional categorization, linkage, resolution, and service fields. Outcome: a new incident is inserted.
1. click('Short description')
2. fill('Short description', <SHORT_DESCRIPTION>)  # e.g. 'Unable to connect to email'
3. click('Caller')
4. fill('Caller', <CALLER>)
5. click('nth=0')  # accept caller suggestion
6. For each additional field, either fill(<FIELD>, <VALUE>), select_option(<FIELD>, <VALUE>), or set_checked(<FIELD>, <VALUE>)  # e.g. 'Impact', 'Category', 'Knowledge'
7. If an additional field is a reference field, click('nth=0') after filling to accept the suggestion  # e.g. 'Resolved by', 'Configuration item', 'Assignment group'
8. click('iframe[name="gsft_main"] >> internal:control=enter-frame >> #sysverb_insert')

Workflow: create_problem
Description: Use when creating a new Problem record. Preconditions: the Problem form is open in create mode. Enter the problem statement, then add optional categorization, service, configuration, and assignment fields. Outcome: a new problem is inserted.
1. click('Problem statement')
2. fill('Problem statement', <PROBLEM_STATEMENT>)  # e.g. 'Oracle Down'
3. For each additional field, either fill(<FIELD>, <VALUE>) or select_option(<FIELD>, <VALUE>)  # e.g. 'Category', 'Service', 'Description'
4. If an additional field is a reference field, click('nth=0') after filling to accept the suggestion  # e.g. 'Assignment group', 'Configuration item'
5. click('iframe[name="gsft_main"] >> internal:control=enter-frame >> #sysverb_insert')

Workflow: apply_list_filter
Description: Use when a ServiceNow list needs to be filtered by one or more conditions joined by AND/OR. Preconditions: the target list is visible. Build the filter row by row, supplying field, operator, and optional typed or selected value for each condition. Outcome: the list displays only records matching the configured filter.
1. click('.list_filter_toggle')
2. select_option(<ROW>, <FIELD>)  # e.g. 'nth=0', 'state'
3. select_option(<ROW>, <OPERATOR>)  # e.g. 'nth=0', '='
4. Set the condition value using either select_option(<ROW>, <VALUE>) or click(<ROW>) then fill(<ROW>, <VALUE>)  # e.g. 'Implement', 'CHG0000007', 'Bow Rugger'
5. To add another condition, click('.filterToolbar .filerTableAction:text-is("<LOGIC>")')  # e.g. 'AND', 'OR'
6. Repeat steps 2-5 for each additional condition
7. click('Run')

Workflow: search_knowledge_base
Description: Use when you need to find information in the knowledge base by querying it
1. fill("Search", "<QUERY>")          # e.g. "\"password to conference room A-561\""
2. click("nth=0")

Workflow: sort_list
Description: Use when a ServiceNow list must be ordered by one or more fields. Preconditions: the target list is visible. Open the list builder, add one sort row per requested field, and set each row's direction. Outcome: the visible records are ordered by the configured sort levels.
1. click('.list_filter_toggle')
2. click('role=button')  # add a sort row
3. select_option(<SORT_ROW>, <FIELD>)  # e.g. 'nth=1', 'Assigned'
4. select_option(<SORT_ROW>, <DIRECTION>)  # e.g. 'nth=1', 'ascending'
5. Repeat steps 2-4 for each additional sort level  # e.g. 'Department', 'descending'
6. click('Run filter')
\end{lstlisting}
\end{example*}

\begin{example*}[Qwen-3.6 Skill Library Learned on WorkArena L2]
\begin{lstlisting}[style=skilldump]

### Skill

1. When using the global navigation filter to open an application, extract the exact application name from the task goal (e.g., 'Cost', 'Organization', 'Change') and fill it into the filter input. Do not guess or default to unrelated modules. Example: for goal 'Filter the organization list by Failed login attempts descending and First name descending.', after click('All'), the correct next action is fill('All menu | placeholder="Filter"', 'Organization') (not fill('All menu | placeholder="Filter"', 'User')).
2. After applying a list filter, always click the first record displayed in the filtered results list to proceed. Do not hallucinate record IDs or click records that do not appear in the current filtered view. Example: for goal 'Update change requests matching "#SERIES-d53387c5-c" in short description with planned start and end dates of 2026-10-23 18:03:56 to 2026-10-24 18:03:56 and 2026-10-24 18:04:56 to 2026-10-25 18:04:56.', after click("Run"), the correct next action is click('Preview record: CHG0047758') (not click('CHG0039945')).
3. During iterative multi-record updates, after saving one record, immediately click the preview card for the next record in the list to continue. Do not navigate back to the list view or home, as this breaks the workflow. Example: for goal 'Update change requests with short description containing #SERIES-2e0ff309-8 to set consecutive planned start and end dates...', after click('iframe[name="gsft_main"] >> internal:control=enter-frame >> #sysverb_update'), the correct next action is click('Preview record: CHG0047728') (not click('Back')).
4. When selecting a record immediately after applying a filter, extract the exact ID from the first visible result card and use it verbatim. Never guess, default to '0000000', or use prefixes not present in the UI. Example: for goal 'Delete all expense line items with short descriptions containing "#d8c6cc4a-8".', after click('Run'), the correct next action is click('Preview record: EXP-0c8e62eac-4') (not click('Preview record: EXP0000000')).
5. Immediately after clicking 'Run' to apply a filter, click the preview card of the first record in the resulting list. Extract the record ID exactly as displayed in the UI without modification or guessing. Example: for goal 'Delete all expense lines with short descriptions containing "#667f4906-3".', after click('Run'), the correct next action is click('Preview record: EXP-1a70c35fb-f') (not click('Preview record: EXP-1874474-4')).
6. During iterative updates, after saving a record, immediately click the preview card of the next record in the list. Do not attempt to interact with form fields or buttons on the saved record. Example: for goal 'Update change requests matching short description LIKE #SERIES-e0c27f06-5 to set planned start and end dates to 2029-06-09 17:43:16-2029-06-12 17:43:16 and 2029-06-12 17:44:16-2029-06-15 17:44:16.', after click('iframe[name="gsft_main"] >> internal:control=enter-frame >> #sysverb_update'), the correct next action is click('Preview record: CHG0040359') (not click('Planned start date | nt
7. Expense-related goals map to the 'Cost' application, not 'Expense'. Extract 'Cost' and fill it into the navigation filter. Example: for goal 'Delete all expense records with a short description containing SERIES-6283ff6a-0.', after click('All'), the correct next action is fill('All menu | placeholder="Filter"', 'Cost') (not fill('All menu | placeholder="Filter"', 'Expense')).
8. During iterative record deletion, after deleting a record, immediately click the preview card of the next record in the list. Do not click 'Yes' or attempt to finish the task until all records are processed. Example: for goal 'Delete all cost expense lines with short descriptions containing "#f219f65c-4".', after click('delete | nth=0'), the correct next action is click('Preview record: EXP-261dffc3c-d') (not click('Yes')).
9. When clicking a record immediately after applying a list filter, always read the actual record ID from the first visible result card in the table. Never construct, guess, or derive the record ID from the filter value or search term. Example: for goal 'Delete all expense records with short descriptions containing SERIES-d75a7d3a-f.', after click('Run'), the correct next action is click('Preview record: EXP-02665716761') (not click('Preview record: EXP-d75a7d3a-f')).

### Workflows

Workflow: apply_filter_and_delete_matches
Description: Filter a list by a specific field and value, then iterate through the results to open and delete each matching record.
1. click('All')          # navigate to application menu
2. fill('All menu | placeholder="Filter"', 'Cost')          # specify target application
3. click('All menu | role=button | Cost')          # open application module
4. click('.list_filter_toggle')          # enable filter row
5. select_option('nth=0', <FIELD>)          # e.g., short_description
6. select_option('nth=0', <OPERATOR>)          # e.g., LIKE
7. click('nth=0')          # focus filter input
8. fill('nth=0', <VALUE>)          # e.g., #SERIES-f258028d-7
9. click('Run')          # execute filter
10. For each matching record <RECORD_ID>:
    - click('Preview record: <RECORD_ID>')
    - click('Open Record')
    - click('delete | nth=0')

Workflow: apply_multi_field_or_filter
Description: Use when records must be retrieved using a complex filter combining multiple fields with `OR` logic. Preconditions: The relevant list menu is accessible. Outcome: The list displays records matching at least one of the specified field/value pairs.
1. click('All')
2. fill('All menu | placeholder="Filter"', <MENU_NAME>)
3. click('All menu | role=button | <MENU_NAME>')
4. click('.list_filter_toggle')
5. select_option('nth=0', <FIELD_1>)
6. select_option('nth=0', <OPERATOR_1>)
7. click('nth=0')
8. fill('nth=0', <VALUE_1>)
9. click('nth=0')
10. click('.filterToolbar .filerTableAction:text-is("OR")')
11. select_option('nth=1', <FIELD_2>)
12. select_option('nth=1', <OPERATOR_2>)
13. click('.filterToolbar .filerTableAction:text-is("OR")')
14. select_option('nth=2', <FIELD_3>)
15. select_option('nth=2', <OPERATOR_3>)
16. click('nth=2')
17. fill('nth=2', <VALUE_3>)
18. click('Run')

Workflow: assign_ticket_by_number
Description: Use when a specific ticket must be located by its exact number and reassigned to a new user. Preconditions: The Service Desk list is visible. Outcome: The ticket's assigned user is updated and saved.
1. click('All')
2. fill('All menu | placeholder="Filter"', <MENU_NAME>)
3. click('All menu | role=button | <MENU_NAME>')
4. click('.list_filter_toggle')
5. click('nth=0')
6. select_option('nth=0', 'number')
7. select_option('nth=0', '=')
8. click('nth=0')
9. fill('nth=0', <TICKET_NUMBER>)
10. click('Run')
11. click('role=button | nth=0')
12. click('Open Record')
13. click('Assigned to | nth=0')
14. fill('Assigned to | nth=0', <ASSIGNEE_NAME>)
15. click('nth=0')
16. click('iframe[name="gsft_main"] >> internal:control=enter-frame >> #sysverb_update')

Workflow: filter_and_update_schedule
Description: When to filter a service list by specific criteria and sequentially update date fields for each resulting record.
1. click('All')          # Navigate to the main module
2. fill('All menu | placeholder="Filter"', 'Change')          # Specify the target application
3. click('All menu | role=button | Change')          # Open the application module
4. click('.list_filter_toggle')          # Open the filter builder
5. for each condition/field:
   select_option('nth=0', '<FIELD>')          # e.g., short_description
   select_option('nth=0', '<OPERATOR>')          # e.g., LIKE
   click('nth=0')          # Focus the condition row
   fill('nth=0', '<VALUE>')          # e.g., #SERIES-123456
6. click('Run')          # Apply the filter to the list
7. for each record in filtered list:
   click('Preview record: <RECORD_ID>')          # e.g., CHG0039945
   click('Open Record')          # Navigate to the detail form
   click('Planned start date | nth=0')          # Select start date input
   fill('Planned start date | nth=0', '<START_DATE>')          # e.g., 2028-12-30 11:23:45
   click('Planned end date | nth=0')          # Select end date input
   fill('Planned end date | nth=0', '<END_DATE>')          # e.g., 2029-01-02 11:23:45
   click('iframe[name="gsft_main"] >> internal:control=enter-frame >> #sysverb_update')          # Save changes

Workflow: apply_list_filter
Description: Apply a single condition filter to a ServiceNow list to retrieve matching records.
1. click('All')          # open global navigation menu
2. fill('All menu | placeholder="Filter"', <APP_NAME>)          # e.g., Cost
3. click('All menu | role=button | <APP_NAME>')          # e.g., Cost
4. click('.list_filter_toggle')          # open filter UI
5. select_option('nth=0', <FIELD>)          # e.g., short_description
6. select_option('nth=0', <OPERATOR>)          # e.g., LIKE
7. click('nth=0')          # focus filter row for input
8. fill('nth=0', <VALUE>)          # e.g., #SERIES-f258028d-7
9. click('Run')          # execute the filter
\end{lstlisting}
\end{example*}

\begin{example*}[GPT-5.4 Skill Library Learned on Mind2Web]
\begin{lstlisting}[style=skilldump]

### Skill

1. When a travel or booking form shows empty labeled fields or collapsed selectors, open the control before typing or choosing values; do not skip directly to filling hidden inputs. Example: for goal 'Find train options from New York, NY - Moynihan Train Hall (NYP) to Washington, DC - Union Station (WAS) for the 15th.', after no prior actions, the correct next action is click("From") (not fill("From", "New York, NY - Moynihan Train Hall (NYP)")).
2. On search/listing pages, satisfy explicit item-condition constraints before secondary perks like shipping, especially when the current results visibly violate the requested condition. Example: for goal 'Find new laptops priced between $400 and $500 with free shipping.', after fill("Search for anything", "new laptop") and click("Search"), the correct next action is click("New") (not click("Free shipping")).
3. Prefer stable navigation or mode controls over content shortcuts when the goal requires changing site context and the page offers a dedicated section/tab for that context. Example: for goal 'Browse museums in New York City.', after no prior actions, the correct next action is click("Cities") (not click("New York City")).
4. When a page shows broad department, section, or mode tabs alongside matching subitems, enter the parent container first before choosing a specific item or filter within it. Example: for goal 'Browse the Wine department with Kroger delivery available by 8:00pm.', after click(""), the correct next action is click("Departments") (not click("Wine")).
5. Prefer click on visible tabs, chips, or pseudo-checkbox labels to activate a filter when the control is presented as a clickable text/value element rather than a standard checkbox input. Example: for goal 'View the Coupons & Deals section filtered to LTO offers.', after click("Coupons & Deals"), the correct next action is click("LTO") (not set_checked("LTO", "true")).
6. When the goal combines a search target with a specific search mode or product type, choose the mode tab that matches the requested service before opening broader category pages or destination-specific shortcuts. Example: for goal 'Book hourly parking at Atlanta International Airport on Spine Road in College Park, Georgia.', after no prior actions, the correct next action is click("Hourly") (not click("Airport")).
7. When the goal names a specific entity but the landing page only shows broad site sections or a search box, start with the site's search control instead of a top-level category or guessed result. Example: for goal 'Save Atlantis to a collection.', after no prior actions, the correct next action is fill("Search", "Atlantis") (not click("Atlantis")).
8. When a date or text field is visibly an input or picker trigger, click it to open/focus the control before typing or selecting a value; do not type into unopened widgets or use select_option on non-dropdown toggles. Example: for goal 'Find hotel options for 1 room for 2 adults and 1 child age 0, staying from the 24th to the 27th, with price range and area filters applied.', after click("hotels."), the correct next action is click("checkin") (not fill("Check In", "24")).
9. After selecting an item in a date-filtered planner, if a visible current date chip/value is shown next to a date picker, click that existing date value to open or change the date instead of typing a new date into the picker label. Example: for goal 'View the planned service changes for Route 4 on Wednesday, May 3, 2023.', after click(""), click("Planned Service Changes"), fill("Search Routes", "4"), and click("4"), the correct next action is click("04/12/2023") (not fill("Select Date", "05/03/2023")).

### Workflows

Workflow: complete_contact_or_request_form
Description: Use when a site presents a lead, inquiry, rental, demo-drive, or checkout-gating form that asks for personal/contact details before continuing. Preconditions: the relevant form is already open. Outcome: required identity/contact fields are filled and the form is submitted or advanced.
1. fill(<NAME_OR_CONTACT_FIELD>, <VALUE>)  # e.g. "First Name", "lastName", "Email Address *", "phone", "zip"
2. fill(<ADDITIONAL_REQUIRED_FIELD>, <VALUE>)  # e.g. address, city, employer, member number, market, message
3. select_option(<OPTION_FIELD>, <OPTION>)  # if present; e.g. state, market
4. click(<SUBMIT_CONTROL>)  # e.g. "Submit", "Submit and Continue", "Sign Up", "Proceed to checkout", "Send Email"

Workflow: search_location_then_choose_suggestion
Description: Use when a site requires typing a place, venue, airport, park, city, destination, or station and then selecting one autocomplete suggestion before proceeding. Preconditions: a searchable location input is visible. Outcome: the intended location is selected into the form or search flow.
1. fill(<LOCATION_FIELD>, <QUERY>)  # e.g. "City or airport", "Where are you going?", "Select a park", "Departure station..."
2. click(<SUGGESTED_LOCATION>)  # e.g. "SFO - San Francisco International, CA", "Alagnak Wild River", "Los Angeles, CA, USA"

Workflow: navigate_to_section_then_open_subpage
Description: Use when you need to reach a topic/category section and then open a related destination within it.
1. click("<SECTION>")          # example: "Music", "News", "Fantasy", "NFL", "Watchlist 5"
2. click("<DESTINATION>")      # example: "Follow", "View All", "Download the App .", "Edit", "VIDEOS", "What to Watch"

Workflow: configure_travel_party
Description: Use when a booking flow asks for traveler or guest counts, room counts, or passenger composition before search. Preconditions: the occupancy/traveler selector is open or accessible. Outcome: the party composition is set.
1. click(<PARTY_CONTROL>)  # e.g. "1 traveler", "1 room, 2 guests", "Number of Guests"
2. click(<INCREMENT_OR_ADD_CONTROL>)  # e.g. "Increase adults", "Add Adults", "Add Rooms"
3. select_option(<AGE_OR_COUNT_FIELD>, <OPTION>)  # if present; e.g. child age, total passengers
4. click(<DONE_CONTROL>)  # e.g. "Done", "Confirm"

Workflow: search_after_form_setup
Description: Use when a site requires entering search criteria and then explicitly launching results. Preconditions: required search inputs have been populated. Outcome: the site navigates to a results page.
1. fill(<SEARCH_FIELD>, <QUERY>)  # e.g. "Search", "Search Site", "Search for anything", "Airport"
2. click(<SEARCH_TRIGGER>)  # e.g. "Search", "Find", "Submit Search", "Get cheapest tickets"

Workflow: browse_category_hierarchy
Description: Use when reaching the target content requires drilling down through two or more navigation levels such as department -> subcategory -> sub-subcategory. Preconditions: top-level navigation is accessible. Outcome: the agent lands on a focused listing or section.
1. click(<TOP_LEVEL_CATEGORY>)  # e.g. "WOMEN", "Sports", "Movies", "Shop"
2. click(<SUBCATEGORY>)  # e.g. "Cropped Tops", "NFL", "In Theaters", "Vitamin C"
3. click(<DEEPER_CATEGORY_OR_SECTION>)  # e.g. "All NFL Tickets", "Coming Soon", "Shirts and Hoodies"

Workflow: apply_filters_to_results
Description: Use when a results or listing page offers one or more facet filters such as size, color, amenities, genre, stock, cancellation, stops, or accessibility. Preconditions: a results page or filter panel is open. Outcome: results are narrowed to match requested constraints.
1. click(<FILTER_GROUP>)  # e.g. "Genre", "Color", "Amenities", "Filter", "Size"
2. click(<FILTER_OPTION>)  # e.g. "comedy", "BLACK", "Business center (11)", "wheelchair"
3. click(<APPLY_OR_SHOW_RESULTS_CONTROL>)  # if present; e.g. "Apply", "Show 10 Results", "Done"

Workflow: sort_results
Description: Use when a listing page supports changing result order after search or filtering. Preconditions: sortable results are visible. Outcome: results are reordered by the chosen criterion.
1. click(<SORT_CONTROL>)  # e.g. "Sort by", "Featured undefined", "Sort by Relevance"
2. click(<SORT_OPTION>)  # or select_option when implemented as a dropdown; e.g. "Price (lowest first)", "Best Sellers undefined"
3. select_option(<SORT_DROPDOWN>, <OPTION>)  # e.g. "Price - Lowest", "Highest Price", "Lowest mileage first"

Workflow: open_item_from_results
Description: Use when the goal is to inspect a specific product, hotel, vehicle, article, game, or media item after search/browse/filtering. Preconditions: a results list is visible. Outcome: the chosen item's detail page or modal is opened.
1. click(<ITEM_RESULT>)  # e.g. "Waking Up: A Guide to Spirituality Without Religion", "Nice Hostel Kameari", "View Details"
2. click(<DETAILS_OR_CONTINUE_CONTROL>)  # if needed; e.g. "See Details", "View Page", "See availability"

Workflow: add_item_to_cart_or_bag
Description: Use when purchasing or collecting a product/add-on requires opening an item and adding it to the cart, bag, or basket. Preconditions: the desired item page or listing action is visible. Outcome: the item is added to the shopping cart/bag.
1. click(<ITEM>)  # e.g. product title, ticket add-on, meal deal
2. click(<ADD_TO_CART_CONTROL>)  # e.g. "Add to cart", "Add To Cart", "Add to Bag", "Add ... to Basket"
3. click(<POST_ADD_CONTINUE_CONTROL>)  # if present; e.g. "No Thanks", "Checkout Now", "View cart"

Workflow: choose_variant_then_add_to_cart
Description: Use when an item must be configured with a size, quantity, or similar variant before adding it to cart. Preconditions: the item detail page is open. Outcome: the selected variant is added to the cart.
1. click(<VARIANT_GROUP>)  # e.g. "Size", quantity selector, ticket quantity
2. click(<VARIANT_OPTION>)  # e.g. "L", "9", "3", "$107 /ea"
3. click(<ADD_TO_CART_CONTROL>)  # e.g. "Add to cart", "Add to Bag"

Workflow: save_item_to_list_or_wishlist
Description: Use when the goal is to save one or more items to a named list, collection, or wishlist rather than purchase immediately. Preconditions: save/list controls are available. Outcome: the item is stored in the chosen list.
1. click(<SAVE_OR_LIST_CONTROL>)  # e.g. "Create a list", "Add To Collection", "Add to Wish List"
2. fill(<LIST_NAME_FIELD>, <LIST_NAME>)  # if creating a new list; e.g. "Add a title (Required)", "Name"
3. click(<CONFIRM_SAVE_CONTROL>)  # e.g. "Create", "Save", "Done", "Save to <LIST>"

Workflow: select_dropdown_options_then_continue
Description: Use when a form or finder is primarily driven by dropdown selections such as year/make/model, make/model, season, state, or fare/class preferences. Preconditions: the relevant select controls are visible. Outcome: the configured options are applied and the flow advances.
1. select_option(<SELECT_FIELD>, <OPTION>)  # e.g. "Year" -> "2006", "Make" -> "Honda", "State" -> "Colorado"
2. select_option(<NEXT_SELECT_FIELD>, <OPTION>)  # e.g. "Model" -> "Civic", season, sort order
3. click(<CONTINUE_OR_SEARCH_CONTROL>)  # e.g. "Next", "Search", "See Cars for Sale"

Workflow: request_booking_or_reservation
Description: Use when, after narrowing results, the site requires a booking/reservation action to proceed to purchase or availability confirmation. Preconditions: a specific result or offering is visible. Outcome: the reservation/booking flow is initiated.
1. click(<RESULT_ACTION>)  # e.g. "See availability", "View Rates", "Check Availability"
2. click(<BOOK_OR_RESERVE_CONTROL>)  # e.g. "Reserve", "Book Now", "Reserve with Genius discount"
3. click(<FINAL_CONFIRM_CONTROL>)  # if present; e.g. "submit", "SELECT"

Workflow: navigate_to_information_section
Description: Use when the task is informational and requires opening a specific tab, subsection, or article within a site after reaching the relevant entity or category. Preconditions: the parent page or section is open. Outcome: the requested informational content is displayed.
1. click(<PARENT_SECTION>)  # e.g. "Stats", "Community", "Find a Park by State", "Support"
2. click(<TARGET_SUBSECTION>)  # e.g. "Season Leaders", "Popular Photos", "Basic Information", "Find My AirPods"
3. click(<REFINEMENT_OR_DATE_CONTROL>)  # if present; e.g. "Next day", "April", "Mar 26"

Workflow: search_then_refine_then_select
Description: Use when a task starts with a keyword search, then applies one or more refinements, then opens a final result. Preconditions: a site search box is available. Outcome: a narrowed result is selected.
1. fill(<SEARCH_FIELD>, <QUERY>)  # e.g. "Search Amazon", "Search Site", "Search for anything"
2. click(<SEARCH_TRIGGER_OR_SUGGESTION>)  # e.g. "Search", autocomplete suggestion
3. click(<FILTER_OR_SORT_OPTION>)  # e.g. "Paperback", "4 Stars & Up", "Price + Shipping: highest first"
4. click(<FINAL_RESULT>)  # e.g. book title, product title, event result

Workflow: navigate_to_category
Description: Use this to open a broad shopping/listing area and then choose a specific category, tab, or subsection.
1. click("<TOP_LEVEL_SECTION>")          # example: "Deals", "Shopping Cart", "Today's Deals", "Coupons & Deals", site/home link
2. click("<CATEGORY_OR_SUBSECTION>")     # example: "DISH Outdoors", "Trending (2)", "LTO", "Inventory", "Shows By Date", "Remove"
\end{lstlisting}
\end{example*}

\end{document}

%% file: math_commands.tex
\usepackage{amsmath,amsfonts,bm}

\def\eqref#1{equation~\ref{#1}}
\def\1{\bm{1}}

\DeclareMathAlphabet{\mathsfit}{\encodingdefault}{\sfdefault}{m}{sl}
\SetMathAlphabet{\mathsfit}{bold}{\encodingdefault}{\sfdefault}{bx}{n}